\documentclass[twoside,twocolumn,9pt]{article}
\usepackage{extsizes}
\usepackage[super,sort&compress,comma]{natbib} 
\usepackage[version=3]{mhchem}
\usepackage[left=1.5cm, right=1.5cm, top=1.785cm, bottom=2.0cm]{geometry}
\usepackage{balance}
\usepackage{sectsty}
\usepackage{graphicx} 
\usepackage{lastpage}
\usepackage[format=plain,justification=justified,singlelinecheck=false,font={stretch=1.125,small,sf},labelfont=bf,labelsep=space]{caption}
\usepackage{float}
\usepackage{fancyhdr}
\usepackage{fnpos}
\usepackage[english]{babel}
\addto{\captionsenglish}{%
  
}
\usepackage{array}
\usepackage{droidsans}
\usepackage{charter}
\usepackage[T1]{fontenc}
\usepackage[usenames,dvipsnames]{xcolor}
\usepackage{setspace}
\usepackage[compact]{titlesec}
\usepackage{hyperref}
\usepackage{algorithm}
\usepackage{algorithmic}
\usepackage{amsmath}
\usepackage{amssymb}
\usepackage[capitalize,noabbrev]{cleveref}
\usepackage{epstopdf}%This line makes .eps figures into .pdf - please comment out if not required.
\definecolor{cream}{RGB}{222,217,201}

\begin{document}

\pagestyle{fancy}
\thispagestyle{plain}
\fancypagestyle{plain}{
%%%HEADER%%%
\renewcommand{\headrulewidth}{0pt}
}
%%%END OF HEADER%%%

%%%PAGE SETUP - Please do not change any commands within this section%%%
\makeFNbottom
\makeatletter
\renewcommand\LARGE{\@setfontsize\LARGE{15pt}{17}}
\renewcommand\Large{\@setfontsize\Large{12pt}{14}}
\renewcommand\large{\@setfontsize\large{10pt}{12}}
\renewcommand\footnotesize{\@setfontsize\footnotesize{7pt}{10}}
\makeatother

\renewcommand{\thefootnote}{\fnsymbol{footnote}}
\renewcommand\footnoterule{\vspace*{1pt}% 
\color{cream}\hrule width 3.5in height 0.4pt \color{black}\vspace*{5pt}} 
\setcounter{secnumdepth}{5}

\makeatletter 
\renewcommand\@biblabel[1]{#1}            
\renewcommand\@makefntext[1]% 
{\noindent\makebox[0pt][r]{\@thefnmark\,}#1}
\makeatother 
\renewcommand{\figurename}{\small{Fig.}~}
\sectionfont{\sffamily\Large}
\subsectionfont{\normalsize}
\subsubsectionfont{\bf}
\setstretch{1.125} %In particular, please do not alter this line.
\setlength{\skip\footins}{0.8cm}
\setlength{\footnotesep}{0.25cm}
\setlength{\jot}{10pt}
\titlespacing*{\section}{0pt}{4pt}{4pt}
\titlespacing*{\subsection}{0pt}{15pt}{1pt}
%%%END OF PAGE SETUP%%%

%%%FOOTER%%%
\fancyfoot{}
%\fancyfoot[LO,RE]{\vspace{-7.1pt}\includegraphics[height=9pt]{head_foot/LF}}
%\fancyfoot[CO]{\vspace{-7.1pt}\hspace{13.2cm}\includegraphics{head_foot/RF}}
%\fancyfoot[CE]{\vspace{-7.2pt}\hspace{-14.2cm}\includegraphics{head_foot/RF}}
\fancyfoot[RO]{\footnotesize{\sffamily{1--\pageref{LastPage} ~\textbar  \hspace{2pt}\thepage}}}
\fancyfoot[LE]{\footnotesize{\sffamily{\thepage~\textbar\hspace{3.45cm} 1--\pageref{LastPage}}}}
\fancyhead{}
\renewcommand{\headrulewidth}{0pt} 
\renewcommand{\footrulewidth}{0pt}
\setlength{\arrayrulewidth}{1pt}
\setlength{\columnsep}{6.5mm}
\setlength\bibsep{1pt}
%%%END OF FOOTER%%%

%%%FIGURE SETUP - please do not change any commands within this section%%%
\makeatletter 
\newlength{\figrulesep} 
\setlength{\figrulesep}{0.5\textfloatsep}

\makeatother
%%%END OF FIGURE SETUP%%%

%%%TITLE, AUTHORS AND ABSTRACT%%%
\twocolumn[
  \begin{@twocolumnfalse}
%{\includegraphics[height=30pt]{head_foot/journal_name}\hfill\raisebox{0pt}[0pt][0pt]{\includegraphics[height=55pt]{head_foot/RSC_LOGO_CMYK}}\\[1ex]
%\includegraphics[width=18.5cm]{head_foot/header_bar}}\par
\vspace{1em}
\sffamily
\begin{tabular}{m{1.5cm} p{14.5cm} }

{}& \noindent\LARGE{\textbf{On-the-Fly Fine-Tuning of Foundational Neural Network Potentials: A Bayesian Neural Network Approach}} \\%Article title goes here instead of the text "This is the title"
\vspace{0.3cm} & \vspace{0.3cm} \\

 & \noindent\large{\;\;\;\;\;\;\;\;\;\;\;\;\;\;\;\;\;\;\;\;\;\;\;\;\;\;\;\;\;\;\;\;\;Tim Rensmeyer, Denis Kramer and Oliver Niggemann} \\ %Author names go here instead of "Full name", etc.
 & \noindent\large{\;\;\;\;\;\;\;\;\;\;\;\;\;\;\;\;\;\;\;\;\;\;\;\;\;\;\;\;\;\;\;\;\;\;\;\;\;\;\;\;\;\;\;\;\;\;Helmut-Schmidt-University Hamburg}\\ \\
{}& \noindent\normalsize{Due to the computational complexity of evaluating interatomic forces from first principles, the creation of interatomic machine learning force fields has become a highly active field of research. However, the generation of training datasets of sufficient size and sample diversity itself comes with a computational burden that can make this approach impractical for modeling rare events or systems with a large configuration space. Fine-tuning foundation models that have been pre-trained on large-scale material or molecular databases offers a promising opportunity to reduce the amount of training data necessary to reach a desired level of accuracy. However, even if this approach requires less training data overall, creating a suitable training dataset can still be a very challenging problem, especially for systems with rare events and for end-users who don't have an extensive background in machine learning. In on-the-fly learning, the creation of a training dataset can be largely automated by using model uncertainty during the simulation to decide if the model is accurate enough or if a structure should be recalculated with classical methods and used to update the model. A key challenge for applying this form of active learning to the fine-tuning of foundation models is how to assess the uncertainty of those models during the fine-tuning process, even though most foundation models lack any form of uncertainty quantification. In this paper, we overcome this challenge by introducing a fine-tuning approach based on Bayesian neural network methods and a subsequent on-the-fly workflow that automatically fine-tunes the model while maintaining a pre-specified accuracy and can detect rare events such as transition states and sample them at an increased rate relative to their occurrence.} \\%The abstract goes here instead of the text "The abstract should be..."

\end{tabular}

 \end{@twocolumnfalse} \vspace{0.6cm}

  ]
%%%END OF TITLE, AUTHORS AND ABSTRACT%%%

%%%FONT SETUP - please do not change any commands within this section
\renewcommand*\rmdefault{bch}\normalfont\upshape
\rmfamily
\section*{}
\vspace{-1cm}
\setcounter{secnumdepth}{2}

\section{Introduction}
Ever since the discovery of the laws of quantum mechanics a century ago, the prediction of molecular and material properties such as stress-strain relationships or catalytic activity from first principles has, in theory, been possible \citep{Atkins, MD_NNP}. However, in practice, this task remains challenging even to this day \citep{MLMQM,MMDFT}. The major difficulty lies in the exponentially growing computational complexity of solving the underlying Schrödinger equation with an increasing number of electrons \citep{Atkins}. As a consequence, several approximate methods for property prediction have been developed, which are computationally tractable for larger systems at the cost of varying degrees of accuracy. \\
Density Functional Theory (DFT) in particular, has established itself as a valuable tool in computational chemistry that allows the computation of many material and molecular properties, such as electronic structure, binding energies and interatomic forces with a high degree of accuracy and a computational complexity that is feasible on a typical high-performance cluster for many tasks \citep{MMDFT}.
While DFT has enabled the investigation of the properties of individual materials at quantum mechanical accuracy, high-throughput screening of materials or molecules for desired properties still remains very computationally demanding.
Furthermore, Molecular Dynamics (MD) -- the simulation of the time evolution of molecular systems and materials -- remains challenging, as the forces on all atoms have to be calculated at each timestep. This severely limits the time horizon that can be achieved in a practical amount of time using DFT \citep{MD_NNP}.\\ 
Subsequently, the development of machine learning models that can predict interatomic forces has become an active field of research \citep{Survey_NNP}. Here, neural networks have become a promising approach. These models have made great strides in the past years and can achieve much higher accuracy than previous methods with just a few hundred well-sampled training configurations of a specific system in some cases \citep{GemNet, SpookyNet, PaiNN, NequIP, NewtonNet, UNiTE}. 
Even more alluring is the prospect of not starting the training from scratch but instead using one of the models from the growing collection of publicly available foundation models. These models have been pre-trained on large databases of materials or molecules as a starting point and fine-tuning such pre-trained models can significantly reduce the amount of training data necessary to reach a desired accuracy when compared to training from scratch \cite{TL1, TL2, TL3, TL4, Artur} .

However, a key challenge that remains is how to select data points for this training dataset.
While existing algorithms for fine-tuning foundation models perform well on benchmark datasets, those benchmark datasets are usually subsampled from very long MD-trajectories in DFT that cover the entire space of atomic arrangements that can occur. This would be way too computationally demanding to generate training data in practice.\\
 At first glance, computing the trajectory at a lower accuracy and computational complexity (e.g., with a minimal basis set) and then recalculating some subset of those configurations at the desired accuracy appears like a simple solution to increase sample diversity and reduce computational demand. However, this is often not feasible. For example, using
 minimal basis sets in DFT can alter the predicted equilibrium bond distances by a few percent when compared to large basis sets. In practice, this can lead to almost disjoint radial distributions as shown for the nitrogen molecule in Figure 1, even though many other physical properties, such as formation energy and vibrational frequency, are similar to higher accuracy methods. 

\begin{figure}
\centering
\includegraphics[scale=0.7]{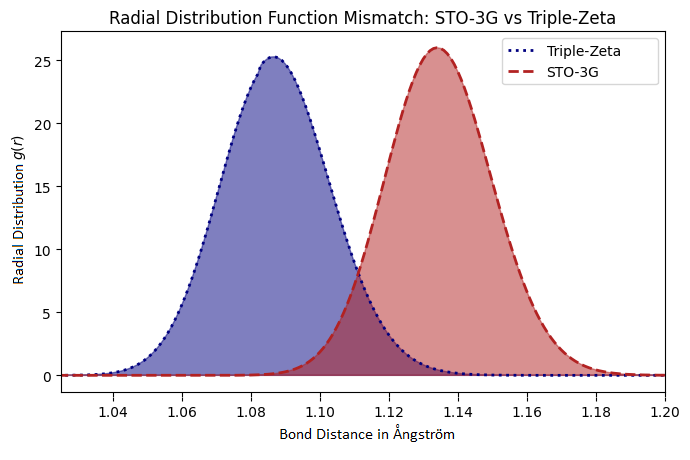}
\caption{Radial distribution functions of a nitrogen molecule at 500 Kelvin computed with a minimal STO-3G basis set and an extensive triple-Zeta basis set. A B3LYP functional \cite{B3LYP1,B3LYP2,B3LYP3,B3LYP4} was used with both basis sets. Because the minimal basis set overestimates the equilibrium bond distance, a simulation at this temperature with this basis set will contain almost no molecular geometries where the bond distance is below the equilibrium bond distance of around $1.085$ \AA \; predicted by the more accurate triple-Zeta basis set}\label{1}
\end{figure}
  
Using the foundational neural network model itself to generate the training data for fine-tuning is also often not that easy. For example, the model might predict the wrong phases at the target temperature, as will be illustrated in a later example. This is especially problematic in instances where the correct phase diagram of the material is not known beforehand.\\
In general, a problem with using a faster approximate method to generate unlabeled training data is the following:
if this method is not accurate enough to generate relevant structures from the process under investigation, then it is not suitable for sampling training data.
However, if it is, then it is questionable if it is even necessary to fine-tune a foundation model since evidently the approximate method is already fast and fairly accurate.\\
Put in another way, many of the application scenarios where fine-tuning models can have the greatest impact are those where there is currently no computationally efficient method that models the system under investigation well, and it is, hence, impractical to collect the training dataset from simulations driven by such methods.

Of course, someone with an extensive background in machine learning force fields might find ways to create a training dataset for some applications. However, the above discussion illustrates the challenges associated with creating such a training dataset. This is especially relevant in the context that the end-users of neural network potentials are generally not researchers on machine learning force fields but computational chemists and material scientists, who should not have to study machine learning force fields in depth to be able to use them.\\
Moreover, for finetuning the foundational neural network potentials inside high-throughput materials discovery workflows, for example, for calculating thermodynamic properties of candidate materials, it would be very desirable if the finetuning process were automated completely without requiring human input for each material.\\

\begin{figure}
\centering
\includegraphics[scale=0.45]{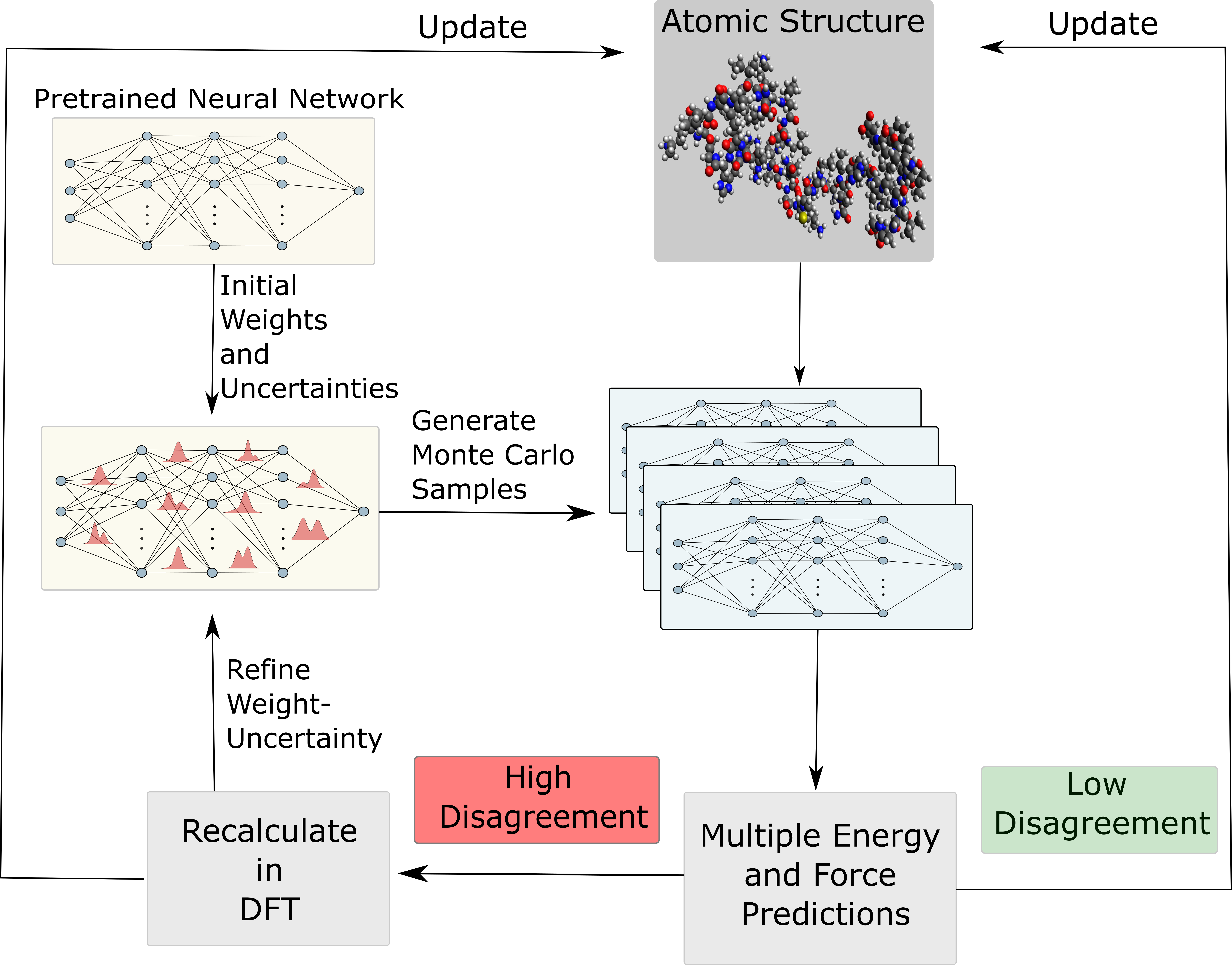}
\caption{An illustration of the on-the-fly learning approach based on Bayesian neural networks introduced in this work.}\label{2}
\end{figure}

Uncertainty-based active learning methods have established themselves as a way to improve data efficiency as well as accuracy during rare events \citep{OTF, AL2, AL3, AL4} by sampling configurations more efficiently. Importantly, this can be achieved in a fairly automated way by selectively labeling samples from regions of configuration space where the MLFF still has a high uncertainty \citep{UGD} and the potential energies are still low enough to be relevant.
Thus, the use of active learning for fine-tuning foundation models appears very promising.\\
For example, on-the-fly learning is an elegant approach \citep{AL1, OTF} where the pre-trained model might be used to drive the dynamics (e.g., MD simulation, transition state optimization, etc.) until a configuration is encountered that exceeds a certain uncertainty threshold. This configuration is then recalculated with classical simulation methods and used to update the model, which then resumes the task until the next configuration above the uncertainty threshold is encountered (Figure \ref{2}).
This approach has the additional benefit of being very easy to use for non-machine learning experts since essentially only the initial state of the system and the uncertainty threshold would have to be specified.\\
Unfortunately, most foundational neural network potentials do not come with an uncertainty estimate for their predictions.
Hence, the development of a framework to systematically update a foundational neural network potential on new data while also being able to assess its uncertainty would be very desirable.\\
Another challenge for fine-tuning foundational neural network potentials on-the-fly is that the fine-tuning algorithm has to be able to fit the training data and avoid overfitting while progressively building up the training dataset from a single initial sample to possibly hundreds of training samples, even though a validation dataset to detect overfitting is unavailable.\\
Moreover, a general challenge in uncertainty quantification with neural networks is that even if the predicted uncertainties are very well correlated with the error, this correspondence is often times not one-to-one. As an example, while a higher standard deviation in the predictions of an ensemble of models might imply a larger error than a smaller one, a standard deviation of $\sigma_{pred}=1$ in the predictions might correspond to an actual standard deviation in the error of $\sigma_{observed}=1.5$. If a validation set exists, the uncertainty estimates can be calibrated. For example, by changing the predicted standard deviation to $\sigma_{pred}\rightarrow \xi\cdot\sigma_{pred}$ where the rescaling factor $\xi$ is estimated on the validation set by matching the mean predicted variance to the observed mean squared error on the validation set. Unfortunately, in the case of on-the-fly learning, this is not an option, since no validation set exists. In general, only the predicted uncertainties and observed errors from the configurations that were recalculated in DFT due to a large uncertainty can be used as empirical data to estimate the miscalibration of the model.\\
A compounding challenge is that if at one point the recalibration factor is wrongfully estimated as too small, this miscalibration might not be correctable, since such a miscalibration inhibits new DFT calls that could be used to calibrate the model more accurately.\\ 

Bayesian Neural Networks have demonstrated a high quality of uncertainty quantification comparable to classical ensemble-based methods of uncertainty quantification for machine learning models of interatomic forces and potential energies \citep{UQEns,rensmeyer}. Further, they are inherently more robust to overfitting by weighing preexisting knowledge in the form of the Bayesian prior density and empirical evidence from training data via Bayes' theorem \citep{SelfSupervisedPrior, SimPrior}.\\
Due to these inherent properties, the principal research question of this paper is whether it is possible to develop a framework for on-the-fly fine-tuning of foundation models based on the formalism of Bayesian neural networks that is capable of addressing the additional challenges mentioned above. 

The main contributions of this work are the following:% \newpage

\begin{itemize}
\item We develop a simple Bayesian framework for uncertainty-aware fine-tuning of foundation models, by harnessing a simple transfer learning prior as well as Monte Carlo Markov Chain (MCMC) sampling of an ensemble of models and assessing the uncertainty via the disagreement in the predictions of the models.
\item We introduce a method for on-the-fly calibration of the uncertainties during the fine-tuning process.
\item We demonstrate that the resulting on-the-fly learning workflow is capable of automatically fine-tuning foundation models to a pre-specified accuracy and biasing the training dataset towards rare events.

\end{itemize}

\section{Interatomic Force Modeling using Bayesian Neural Networks}
Throughout the rest of this paper, bold case symbols designate vectors and $\boldsymbol{y}|\boldsymbol{x}$ denotes $\boldsymbol{y}$ conditioned on $\boldsymbol{x}$.\\ 
Machine-learned interatomic force fields for molecules aim to map an atomic configuration $\{(\boldsymbol{r}_1, z_1), ...,(\boldsymbol{r}_n, z_n) \}$, composed of the nuclear coordinates $\boldsymbol{r}_i \in \mathbb{R}^3$ and nuclear charges $z_i$, to the potential energy $E$ and forces $\boldsymbol{F}_i$ acting on each nucleus $i \in \{1, ..., n\}$. The predicted forces can then, for example, be used in combination with Newton's equations of motion to model the time evolution of the molecules. For materials, the input will usually also contain the lattice vectors, and the stress tensor will often be predicted as well.
Because generating large amounts of high-quality training data is typically infeasible due to the high computational demand of classical simulation methods, modern machine learning models have several forms of physical constraints built into them, in order to make them more data efficient \citep{ SpookyNet, PaiNN,  NequIP, UNiTE, SchNet, Conservation}, such as energy conservation by calculating the forces as the negative analytical gradient of the potential energy with respect to the atomic positions. Further, rotation invariance of the potential energy is explicitly built into modern neural network architectures.

\subsection{Bayesian Neural Networks}
Bayesian neural networks have demonstrated promising results for modeling uncertainties in neural network predictions and, in particular, in machine learning force fields \citep{UQEns, BNN_quality, rensmeyer}.
The main difference between the Bayesian approach to neural networks and the regular approach is that the trainable parameters of the neural network, e.g., its weights and biases, are modeled probabilistically. For simplicity of notation, we denote by $\boldsymbol{\theta}$ a vector containing all the trainable parameters of the neural network. For a given parameter vector $\boldsymbol{\theta}$ and input sample $\boldsymbol{x}$, the neural network predicts a probability density $p(\boldsymbol{y}|\boldsymbol{x},\boldsymbol{\theta})$ over the target variable $\boldsymbol{y}$. In the case of machine learning force fields for non-periodic systems $\boldsymbol{x}$ will be an atomic configuration $\{(\boldsymbol{r}_1, z_1), ...,(\boldsymbol{r}_n, z_n) \}$  and $\boldsymbol{y}$ will be the potential energy and atomic forces $\{E,\boldsymbol{F}_1, ..., \boldsymbol{F}_n\}$. For periodic systems, $\boldsymbol{x}$ will contain the lattice vectors as well, and $\boldsymbol{y}$ might additionally contain the stress tensor.
The starting point of Bayesian methods is a prior density $p(\boldsymbol{\theta})$ over the parameters, which expresses a priori knowledge about which sets of parameters are likely to result in a good model of the underlying data distribution. Given some training dataset $\mathcal{D}=\{(\boldsymbol{x}_1,\boldsymbol{y}_1), ..., (\boldsymbol{x}_m,\boldsymbol{y}_m)\}$ the prior density gets refined into the posterior density $p(\boldsymbol{\theta}|\mathcal{D})$ using Bayes' theorem: \[p(\boldsymbol{\theta}|\mathcal{D})=\frac{p(\mathcal{D}|\boldsymbol{\theta})p(\boldsymbol{\theta})}{p(\mathcal{D})}.\] With the mild assumption on conditional independence \[p(\boldsymbol{y}_1, ..., \boldsymbol{y}_m|\boldsymbol{x}_1, ..., \boldsymbol{x}_m, \boldsymbol{\theta})=\Pi_{i=i}^m p(\boldsymbol{y}_i|\boldsymbol{x}_i,\boldsymbol{\theta})\] this can be simplified to
 \[p(\boldsymbol{\theta}|\mathcal{D})=Z\cdot p(\boldsymbol{\theta})\Pi_{i=i}^m p(\boldsymbol{y}_i|\boldsymbol{x}_i,\boldsymbol{\theta}),\] where $Z=\frac{p(\boldsymbol{x}_1, ..., \boldsymbol{x}_m)}{p(\mathcal{D})}$ is a normalization constant.
On a new input sample $\boldsymbol{x}$, the probability distribution of the target variable $\boldsymbol{y}$ can then be calculated via \[p(\boldsymbol{y}|\boldsymbol{x},\mathcal{D})=\int p(\boldsymbol{y}|\boldsymbol{x},\boldsymbol{\theta})p(\boldsymbol{\theta}|\mathcal{D})d\boldsymbol{\theta} =\mathbb{E}_{p(\boldsymbol{\theta}|\mathcal{D})}\left[ p(\boldsymbol{y}|\boldsymbol{x},\boldsymbol{\theta})\right].\]
Because this integral is almost never analytically tractable for neural networks, a Monte Carlo estimate is typically used: \[ p(\boldsymbol{y}|\boldsymbol{x}, \mathcal{D})\approx \frac{1}{k}\sum_{i=1}^{k}p(\boldsymbol{y}|\boldsymbol{x},\boldsymbol{\theta}_i), \;\;\boldsymbol{\theta}_i \sim p(\boldsymbol{\theta}|\mathcal{D}), \]
where the parameter sets $\boldsymbol{\theta}_i$ are sampled from the posterior density using either approximate inference \citep{Swag, BBB, BNNP_DO} or MCMC methods \citep{SGLD,SGHMC, Complete_Recipe}. MCMC methods, in particular, have displayed good results in uncertainty quantification \citep{BNN_quality1} due to their ability to sample different regions of the posterior. These methods work by simulating a stochastic process over the space of neural network parameters, which converges in distribution to the posterior.

\subsection{Relation to other Works}
Even though Bayesian neural networks offer a very promising opportunity to systematically incorporate and update pre-existing knowledge via the prior density, we find that this approach is very underutilized in the literature. In fact, we could only find two instances in the literature where this was attempted \citep{SelfSupervisedPrior, SimPrior}.
In the work by \citet{SimPrior}, transfer learning of simulated to experimental data was done via a Bayesian neural network prior. A simple isotropic Gaussian prior with a mean derived during pretraining was used in that work, which we will also employ here.\\
A more sophisticated approach for constructing a transfer learning prior was introduced by \citet{SelfSupervisedPrior}, where a rescaled local approximation of the posterior on the pretraining dataset was used as a prior. However, in the applications considered here, this approach is not practical since such a prior would have to be constructed during the pretraining of the foundation models, while we focus on fine-tuning foundation models that have already been pre-trained.\\
Transfer learning of a pre-trained model for interatomic force fields has been investigated by \citet{TL1, TL2, TL3} and \citet{ TL4}. However, the modeling of uncertainty has not been under consideration in those works. \\
After presenting our initial investigation of the uncertainty-aware transfer learning approach in a non-archival peer-reviewed venue \citep{BTL}, there has very recently been one other work published that does uncertainty-aware fine-tuning \cite{UTL}. However, the integration into an active learning workflow, like on-the-fly learning, is not under investigation in that work. Further, they investigate instances where large training datasets of thousands of training samples are available and are more focused on fine-tuning for entire subclasses of materials. In contrast, we focus on fine-tuning for specific systems where the size of the training dataset varies from one to a few hundred training samples.\\
Finally, there are the seminal papers by \citet{OTF2} and \citet{OTF} for on-the-fly learning for molecular dynamics. However, they use Gaussian processes as their substitutional model, which are less data efficient than modern neural network architectures and furthermore require training from scratch instead of fine-tuning a pre-trained model.

\section{Methodology}
The fundamental prior assumption for fine-tuning neural network models to specific applications is that the weights of the pre-trained model are not quite right for the application and have to be adjusted by an unknown small change. The idea for the Bayesian fine-tuning approach is, that we model this uncertainty in the correct weights explicitly via a transfer learning prior that is a Gaussian distribution over the weights $p_{TL} \sim N(\boldsymbol{\theta}_{0},\sigma_{TL}^2 I)$ which is centered around the weights of the pre-trained model $\boldsymbol{\theta}_{0}$ and has a small standard deviation $\sigma_{TL}^2$. As we progressively build up our training dataset, we refine the initial uncertainty in the weights by applying Bayes' rule to calculate the posterior.
From the posterior, we then generate several sets of neural network weights, resulting in an ensemble of models. We can then assess the uncertainty in the prediction for new samples via the disagreement between the individual models.\\

To calibrate the uncertainties on the fly, we use the following Bayesian procedure: \\
The uncalibrated Bayesian Neural Network (BNN) models the error $e_{obs}=E_{pred}-E_{true}$ as $$p(e|\sigma_{pred}):=p(e_{obs}=e|\sigma_{pred})=\frac{1}{\sqrt{2\pi\sigma_{pred}^2}}\exp\left(-0.5\frac{e^2}{\sigma_{pred}^2}\right).$$
Here, the predicted energy is the mean predicted energy of all Monte Carlo samples and $\sigma_{pred}$ is calculated from their disagreement (see Appendix B.1-4 and B.9 for details).
Similar to what we described in the introduction, we want to calibrate this distribution by introducing the calibration parameter $\lambda$ to model the error distribution as 
$$p(e|\sigma_{pred},\lambda)=\frac{\sqrt{\lambda}}{\sqrt{2\pi\sigma_{pred}^2}}\exp\left(-0.5\lambda\frac{e^2}{\sigma_{pred}^2}\right).$$
To calibrate the model, we want to estimate a suitable value for $\lambda$. To do this, we use a Bayesian estimator for $\lambda$ by introducing a prior in the form of a Gamma distribution over $\lambda$:
$$p(\lambda)=Gam(\lambda|a,b)=\frac{b^a\lambda^{a-1}}{\Gamma(a)}\exp(-b\lambda).$$
Here, $\Gamma$ denotes the gamma function.
Doing this allows us to utilize pre-existing knowledge from previous experiments about what values are more likely for $\lambda$. Furthermore, it enables us to bias the calibration towards larger uncertainties during the beginning of the run, where there is not enough empirical data to accurately estimate $\lambda$ and thus avoid mistakenly making the model incorrigibly overconfident by approaching the correct calibration from the direction of underconfidence with a growing dataset. Lastly, because Gamma distributions are conjugate priors for Gaussians, using this form of a prior and following the Bayesian procedure makes it possible to derive analytical expressions for the probability that the magnitude of the error is smaller than a predefined threshold $K>0$ as 
$$p(|e^*|<K|\sigma^*,E,\Sigma)=\frac{2K\Gamma(a+\frac{n+1}{2})}{\sqrt{2\pi\sigma^{*2}}\Gamma(a+\frac{n}{2})\sqrt{b+\frac{1}{2}n\cdot M_n}}$$
$$\;\;\;\;\;\;\; \times Hyp2F1\left(\frac{1}{2},a+\frac{n+1}{2};\frac{3}{2},-\frac{K^2}{\sigma^{*2}(2b+n\cdot M_n)}\right),$$
 where $Hyp2F1$ denotes the hypergeometric function 2F1 and $M_n=\frac{1}{n}\sum_{i=1}^n\frac{e_i^2}{\sigma_i^2}$ (see Appendix B.11 for more details of the derivation for this result).\\

We use this result to decide if a DFT calculation should be done by specifying an error threshold $K$ and calling a DFT calculation if the predicted probability of the error magnitude exceeding $K$ is larger than five percent.\\
If the probability is smaller than five percent, the model continues with the MD-simulation. Otherwise, a DFT reference calculation is done and added to the training dataset. Afterwards, we then use Markov chain sampling to generate an ensemble of eight neural networks from the updated posterior.
To sample the posterior density, we use the AMSGrad version of the Stochastic Gradient Hamiltonian Monte Carlo (SGHMC) algorithm \citep{SGHMC} introduced by us in a previous work \citep{rensmeyer}. We use the old Monte Carlo samples as a starting point for sampling the new ones and simulate short Markov Chains with those initial seeds and priority sampling of the newly added structure to achieve faster convergence (see Appendix B.5 for more details). 
Finally, the DFT forces and stresses are used to perform this MD step. 
The resulting workflow is illustrated in Figure \ref{2}.\\

For our experiments, we use three different neural network potentials, NequIP\cite{NequIP}, MACE\cite{MACE, MACE2} and Equiformerv2\cite{equiformerv2}. To apply the Bayesian neural network formalism to these models, we add a few layers to the models so that each model predicts a distribution instead of making point predictions. The details of these modifications, as well as additional details about how predictions are done with the BNN, can be found in Appendix B.\\

We always label the initial atomic structure with DFT and include this first data point to sample the initial Monte Carlo samples. Because potential energies are only defined up to a constant and this constant might be different for the data the model was pre-trained on and the simulation software used for fine-tuning, we also use this initial labeled data point to construct an offset value that is added to all energies calculated by DFT during the on-the-fly simulation. This value is chosen so that the offset DFT calculated energy equals the potential energy predicted by the pre-trained model on the initial structure.

\section{Empirical Evaluation}
The promise of transfer learning is to specialize a pre-trained model with limited computational effort to truthfully reflect subtle, but important aspects of a particular system of interest that are not adequately reproduced by the starting model. Hence, we have selected an initial illustrating example from a classical benchmark dataset of a surface-adsorbate system for neural network potentials, showcasing the improvement in data efficiency and the quality of the uncertainty quantification that results from the Bayesian neural network priors constructed from a pre-trained model.
We then selected a number of “challenging” cases to demonstrate the power and general utility of the full on-the-fly finetuning workflow: we investigate a simple molecular system with several meta-stable conformers to highlight the ability to bias the training data towards transition states between meta-stable states, before turning to two solid state problems; first, we investigate a challenging phase transition due to subtle electronic effects in a highly relevant compound; finally, we address a dynamic problem where mobility of a migrating species is highly dependent on the dynamics of the host lattice.\\

To illustrate the effect of training a Bayesian neural network model with the prior constructed from the weights of a pre-trained model, compared to training a model from scratch, we start out with a transfer learning scenario for potential energies of a surface adsorbate system. We use a publicly available EquiformerV2 model \cite{equiformerv2} which was pre-trained on the OC20 dataset \citep{oc20}. For the target dataset, we choose the NbSiAs surface - COH adsorbate dataset from the OC20-Dense dataset \citep{oc20dense}. Notably, this benchmark does not yet involve the on-the-fly learning workflow and is instead done on a publicly available benchmark dataset. We partition the dataset into training, validation, and test data and then generate Monte Carlo samples from the posterior resulting from the training dataset (see Appendix B.5 and C.4 for details). Here we compare the model accuracy to a comparable BNN that is trained with an uninformative prior. Further, we illustrate the quality of the resulting uncertainty quantification by calculating the Area Under the Curve - Receiver Operating Characteristic (AUC-ROC) scores for determining whether a test sample has a prediction error that lies above or below 1 kcal/mol based on the standard deviation in the predictions of the model. Because recognizing samples with an error above a certain threshold becomes easier at higher accuracies, we plot the AUC-ROC scores over the Root Mean Square Errors (RMSEs) on the test set for this metric. To assess the viability of this transfer learning approach to finetuning models, we initially also did additional transfer learning scenarios, which were focused on transfer learning of molecular forces with a NequIP model in several settings. These experiments and their results are listed in Appendix A.\\

As the first benchmark for the proposed on-the-fly fine-tuning workflow, we investigate the on-the-fly fine-tuning of a NequIP \cite{NequIP} model during a molecular dynamics simulation of an ethanol molecule at 300 Kelvin. The NequIP model was pre-trained by us on the SPICE dataset. Despite its small size, ethanol is an interesting benchmark molecule because it has different meta-stable conformations - one anti and two symmetry-related gauche conformations (Figure \ref{3}). Due to this, there are transition states between those conformations that are expected to occur during an MD-simulation at this temperature, which are relatively rare due to the higher potential energy when compared with the metastable conformations. In a good on-the-fly learning workflow, such transition states have to be identified by the uncertainty measure and added to the training dataset at a higher rate compared to the rate of their occurrence during the simulation to ensure the accuracy of the model during transitions. For this benchmark, we set the target accuracy as 0.5 kcal/mol and perform a 10 ps simulation.\\

For the second experiment with the proposed on-the-fly fine-tuning algorithm, we run a simulation of a 2x2x2 LaMnO\textsubscript{3} supercell with the mace-mp0 medium model \cite{MACE,MACE2}. This system is interesting because it undergoes a phase transition at around 750 Kelvin (Figure \ref{7} a). LaMnO\textsubscript{3} crystallizes in an orthorhombic phase with space group Pbnm below 750K due to the strong Jahn-Teller activity of manganese. The orthorhombic phase is an anti-ferromagnetic insulator in which an orbital ordering is established due to the cooperative Jahn-Teller effect breaking the degeneracy of the electronic configuration of  Mn\textsuperscript{3+}$(t^{3}_{2g}e^{1}_{g})$\cite{LaMnRef}.
For this reason, we perform a 150-ps molecular dynamics simulation with two temperature jumps. For the first 25 ps, we run the simulation at 300 Kelvin. At the 25 ps mark, we introduce a temperature jump to 800 Kelvin. We keep this temperature for 100 ps after which we introduce a second temperature jump back to 300 Kelvin and continue the simulation for another 25 ps.\\

The final benchmark for the proposed on-the-fly fine-tuning algorithm is a proton diffusion simulation in a 160-atom CaZrS\textsubscript{3} supercell at 1500 Kelvin, again with the mace-mp0 medium model. This benchmark is challenging because proton mobility is intimately linked to the dynamics of the host lattice. Occasional close proximity of two sulfur atoms belonging to non-connected ZrS\textsubscript{6} octahedra enables a direct jump between the two ZrS\textsubscript{6} octahedra with virtually zero activation energy \cite{Stefan} (Figure \ref{8} a), resulting in unusually high proton mobility at low temperatures. The S-S distance is governed by a static lattice distortion due to size mismatches between Ca and Zr as well as the vibrational properties of the sulfur sublattice. Hence, this benchmark goes beyond the mere reflection of transition state energies. It also requires a correct representation of the dynamics of the host lattice. Accurately modeling the proton transport in this system, therefore, might require a very high accuracy of the model. Here we evaluate the model with two different error thresholds, 5 kcal/mol and 15 kcal/mol. We fine-tune both models on-the-fly during an initial MD simulation and then use the fine-tuned models and the non-fine-tuned one to investigate the diffusion of protons in CaZrS\textsubscript{3}.\\

All experiments were done with 8 Monte Carlo samples, which we identified as a good tradeoff between computational complexity and quality of uncertainty quantification in previous benchmarks \citep{rensmeyer}. Details of the sampling procedures can be found in Appendix B.5. More details for all the experiments can be found in Appendix B. We set $\sigma_{TL}$ from the Bayesian neural network prior as $0.2$ for all experiments involving the NequIP model, $0.5$ for the MACE model and $0.02$ for the EquiformerV2 model. Based on the results of our initial transfer learning experiments, we chose a=3 and b=10 for the Gamma-prior over the calibration parameter for the NequIP model. Because we did not have such results for the MACE model, we chose the more conservative parameters a=1.5 and b=10 for the corresponding experiments.

\section{Results}
\subsection{Efficiency of the Transfer Learning Approach on the NbSiAs Surface - COH Adsorbate Dataset}
\begin{figure}%[h]%[H]
%\centering
\includegraphics[scale=0.95]{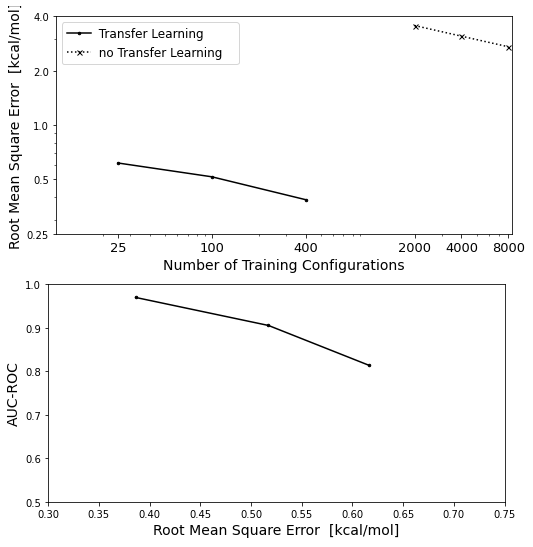}
\caption{Results of the transfer learning approach on the NbSiAs surface - COH adsorbate dataset. Transfer learning via the prior constructed from the pre-trained model leads to a significant improvement in data efficiency (upper graph). Further, the predicted uncertainties are well suited for identifying which test samples have an error above 1 kcal/mol (lower graph). }\label{4}
\end{figure}
Plotting the root mean square errors of the models for the potential energies on the test set, we see a considerable improvement in accuracy at a given size of the training dataset (Figure \ref{4} upper graph).
Furthermore, the AUC-ROC scores for distinguishing low and high error samples from the test set based on the predicted uncertainty, which are shown in the lower graph of Figure \ref{4}, demonstrate that the predicted uncertainty of the fine-tuned model is well suited for distinguishing low and high error samples. Also, we see the improvement in AUC-ROC scores on this benchmark with decreasing root mean square error on the test set, which was expected because samples with an error of more than 1 kcal/mol become more extreme outliers with increasing accuracy of the model.
 Note that an AUC-ROC score of 1 corresponds to perfect distinguishability and an AUC-ROC score of 0.5 corresponds to random guessing. All Monte Carlo samples were generated from a single very long Markov Chain, to make sure no pathological overfitting occurs (see Appendix B.5 for more details).

\begin{figure}[h!]
%\centering
\includegraphics[scale=0.1]{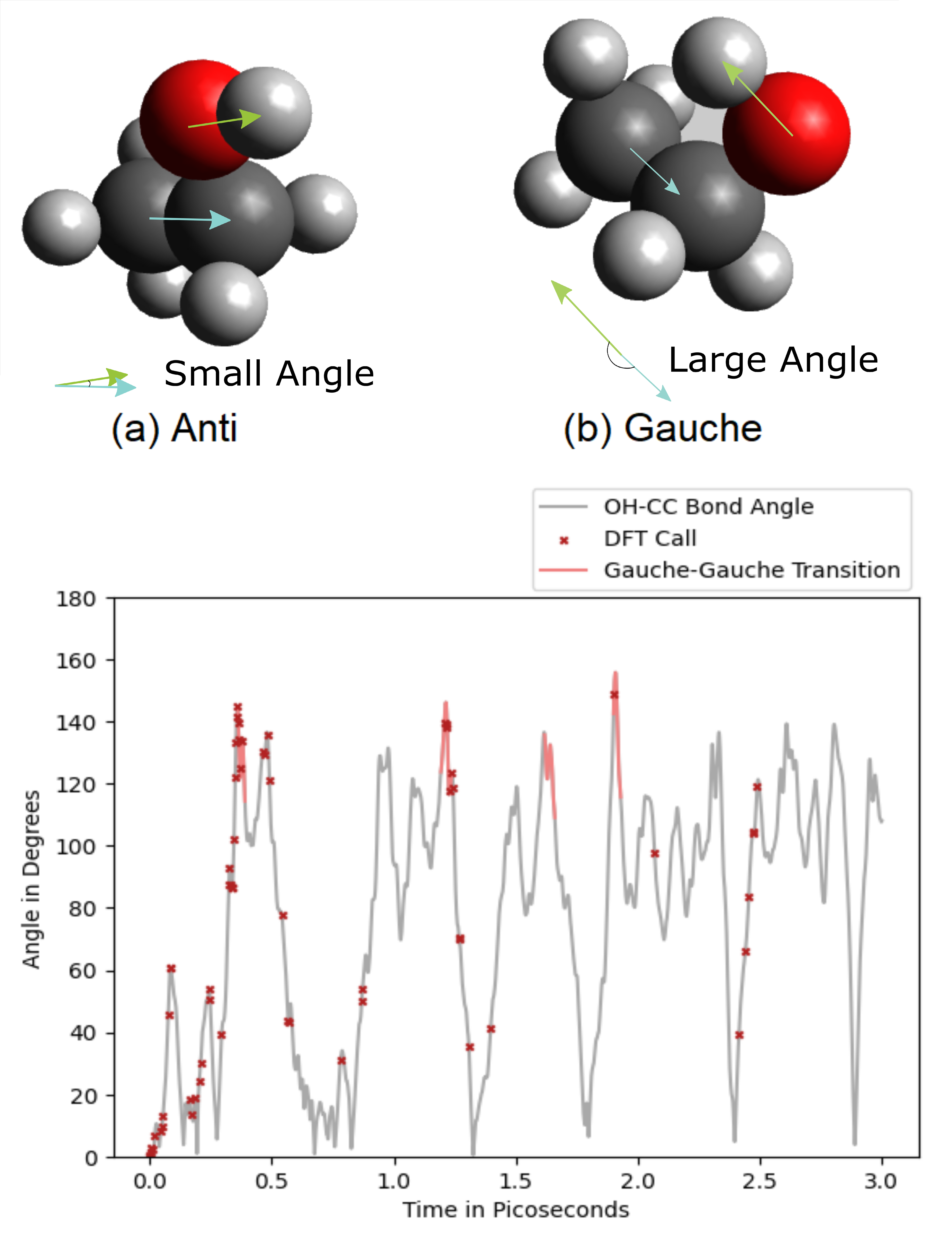}
\caption{Upper Figure: The meta-stable conformations of Ethanol.\\
%The low and high-temperature phases of LaMnO3. On the left, the low-temperature phase with its characteristic tilted manganese oxygen polyhedron with 3 distinct Mn-O bond lengths. On the right, the more symmetric high-temperature phase with all Mn-O bond lengths being roughly equal.\\
Lower Figure:  Angle between the OH and CC bond axis of the ethanol molecule over time. A small angle corresponds to the anti-conformation and a large angle to the gauche-conformation. The gauche-gauche transitions are colored since they are not well identifiable by the angle. After the initial training phase, almost all DFT calls happen during transition states. }\label{3}
\end{figure} 
\subsection{Conformers and Transition States of Ethanol}

On the ethanol benchmark, we find a very good consistency of the proposed on-the-fly fine-tuning algorithm with the specified error threshold of 0.5 kcal/mol. Overall, we find that for the 65 DFT interventions, the model exceeded the error threshold 3 times, corresponding to 4.55 percent of interventions with an overall RMSE of 0.196 kcal/mol at the interventions. Only two DFT references were called after the first 5 ps of the simulation. We performed an additional analysis of 150 structures randomly sampled from the 10 ps trajectory, where the model exceeded the threshold only a single time, with an overall RMSE of 0.142 kcal/mol, significantly outperforming the non-fine-tuned model, which has an RMSE of 0.958 kcal/mol on those structures. \\
To evaluate this experiment qualitatively, we also analyzed the DFT interventions in relation to transition states between different meta-stable conformations. For this, we plotted the angle between the OH bond and the CC bond over time. By doing this, the anti-conformation can be identified by a small angle and the gauche conformation by a large angle. Further, transitions between anti and gauche conformations can be identified by large changes in the angle. Since gauche-gauche transitions are not very well visible via this angle, we highlighted them manually by labeling them by hand. The results are summarized in Figure \ref{3}. As can be seen in this figure, after the initial learning phase, where the model has high uncertainty in general, the model is almost exclusively requesting DFT calls during the transitions between meta-stable conformations. Especially noteworthy is that in the initial three-ps timeframe, where almost all DFT calls happen, the Gauche-Gauche transitions make up only around five percent (0.155 ps) of the total time, but around 16 percent (10 out of 63) of the total DFT interventions of that timeframe happen during those transitions. Hence, Gauche-Gauche transition states were sampled at a more than three times higher rate relative to their rate of occurence.

\subsection{Phase Transition in LaMnO\textsubscript{3}}

 \begin{figure*}
\centering
\includegraphics[scale=0.4]{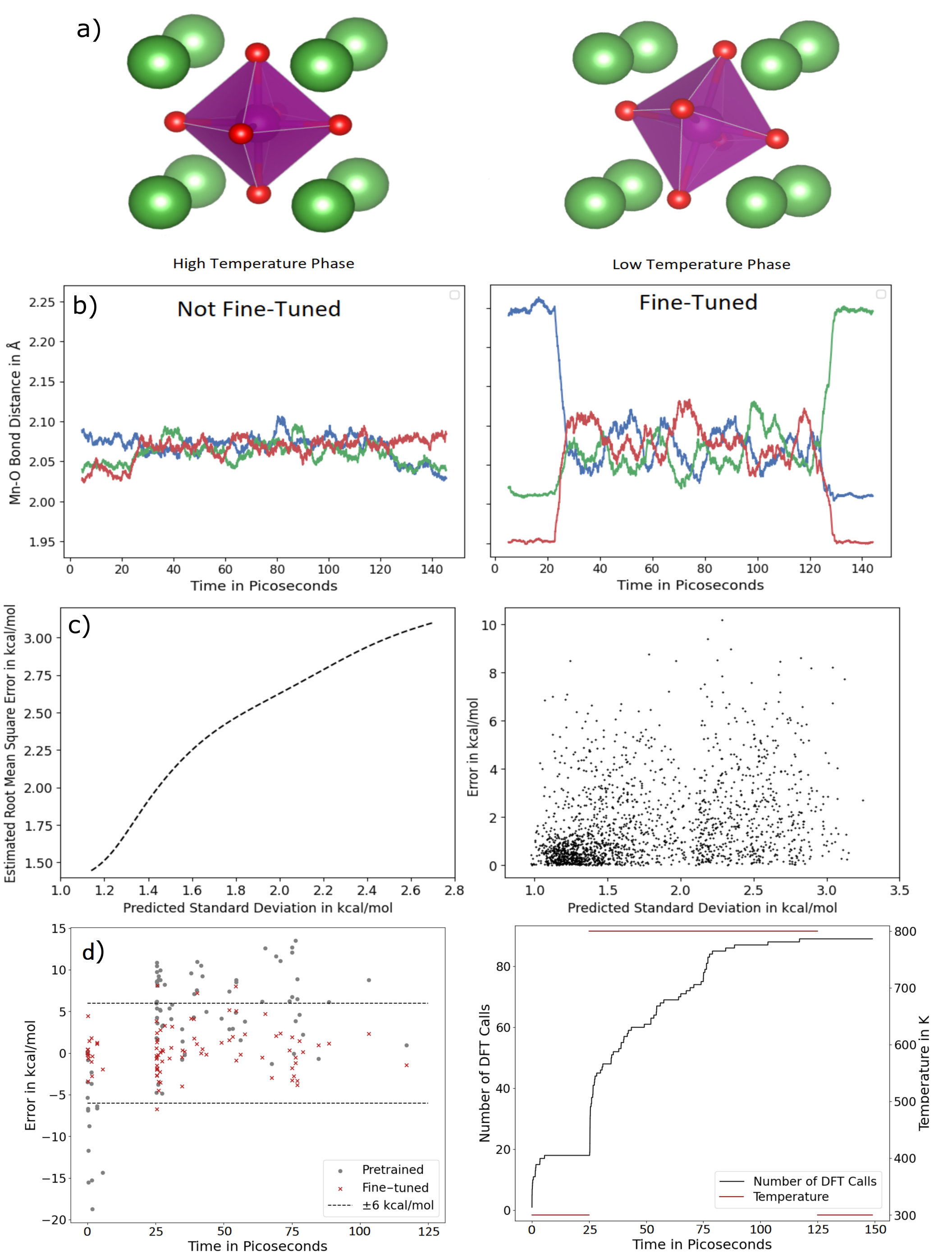}
\caption{a)The low and high-temperature phases of LaMnO\textsubscript{3}. On the left, the more symmetric high-temperature phase with all Mn-O bond lengths being roughly equal. On the right, the low-temperature phase with its characteristic tilted manganese oxygen polyhedron with 3 distinct Mn-O bond lengths. \\
b) Five picosecond moving averages for the Mn-O bond distances over time. On the left, the non-fine-tuned model. On the right, our on-the-fly fine-tuned model. The non-fine-tuned model wrongfully predicts the high-temperature phase during the low-temperature intervals of the simulations, as can be seen by the roughly equal Mn-O bond distances. In contrast, the fine-tuned model predicts the correct phase behavior, with the low-temperature phase, with the three distinct Mn-O bond lengths being present at the low-temperature intervals of the simulation.\\
c) Predicted standard deviation and error on the LaMnO\textsubscript{3} system. The RMSE increases well with increasing standard deviation. However, some instances of structures with low predicted standard deviation and large error still exist. The standard deviations have been calibrated with the maximum likelihood a-posteriori value of the posterior over the rescaling parameter evaluated at the end of the simulation. \\
d)DFT calls over time on the LaMnO\textsubscript{3} system. On the left, the error of the model during DFT reference calls compared to the error of the non-fine-tuned model for those structures. The dashed lines represent the specified 95 percent confidence region for the error of the model during interventions.}\label{7}
\end{figure*}

For the LaMnO\textsubscript{3} system, we also find good consistency with the error threshold, with 4.5 percent of configurations at intervention time having an error above the threshold of 6 kcal/mol. At the time of intervention, the on-the-fly learning model had an RMSE of 2.685 kcal/mol compared to an error of the non-fine-tuned model of 7.532 kcal/mol on the same structures. In contrast to the non-fine-tuned model, the fine-tuned model reproduces the correct phases at the two different temperatures, as can be seen by the three distinct MnO bond distances at lower temperatures in Figure \ref{7} b. The observed bond distances of the fine-tuned model are in close agreement with the values of 2.235 \AA, 2.002 \AA \;and 1.947 \AA\; that were computed by \citet{LaMnO3Ref} with DFT at the PBE+U level of theory, also used for this benchmark. To assess the quality of the uncertainty quantification, we recalculated 1800 structures from the simulation in DFT and evaluated the relationship between predicted (calibrated) standard deviations and error by scatter plotting the uncertainties over the error magnitude (Figure \ref{7} c). Further, we estimated the RMSE at a given standard deviation by ordering the pairs of errors and predicted standard deviations by order of ascending standard deviations and then applying Gaussian smoothing (see Appendix B.10 for details). As can be seen in this figure, we have an overall good relationship between predicted standard deviation and error. However, it should be noted that there are instances where the error is quite large despite a low predicted standard deviation. Note that in order to better estimate the error at higher uncertainties, we did additional sampling of structures with a predicted standard deviation of around 2.1 kcal/mol, 2.5 kcal/mol and 2.8 kcal/mol respectively, which causes the increase in the point density of the scatter plot at these values. Evaluating the accuracy for the sampled structures, we find that the model achieves an RMSE of 2.350 kcal/mol compared to 10.827 kcal/mol of the non-fine-tuned model, with 3.2 percent of samples being above the error threshold.\\
Lastly, we also evaluated the number of DFT calls over time (Figure \ref{7} d). We find that the model stops doing DFT reference calls very early during the initial low-temperature phase of the simulation. After the temperature jump, it starts doing DFT calls again, with the frequency decreasing over time. Further, the model stays well within the specified accuracy region with only four instances where the error is larger than the threshold at the time of intervention. For reference, we also plot the error of the non-fine-tuned model on the structures for which DFT calls were made, illustrating the improvement in accuracy.
%/newpage
\subsection{Proton Mobility in CaZrS\textsubscript{3}}
\begin{figure*}%[H]
\centering
\includegraphics[scale=0.97]{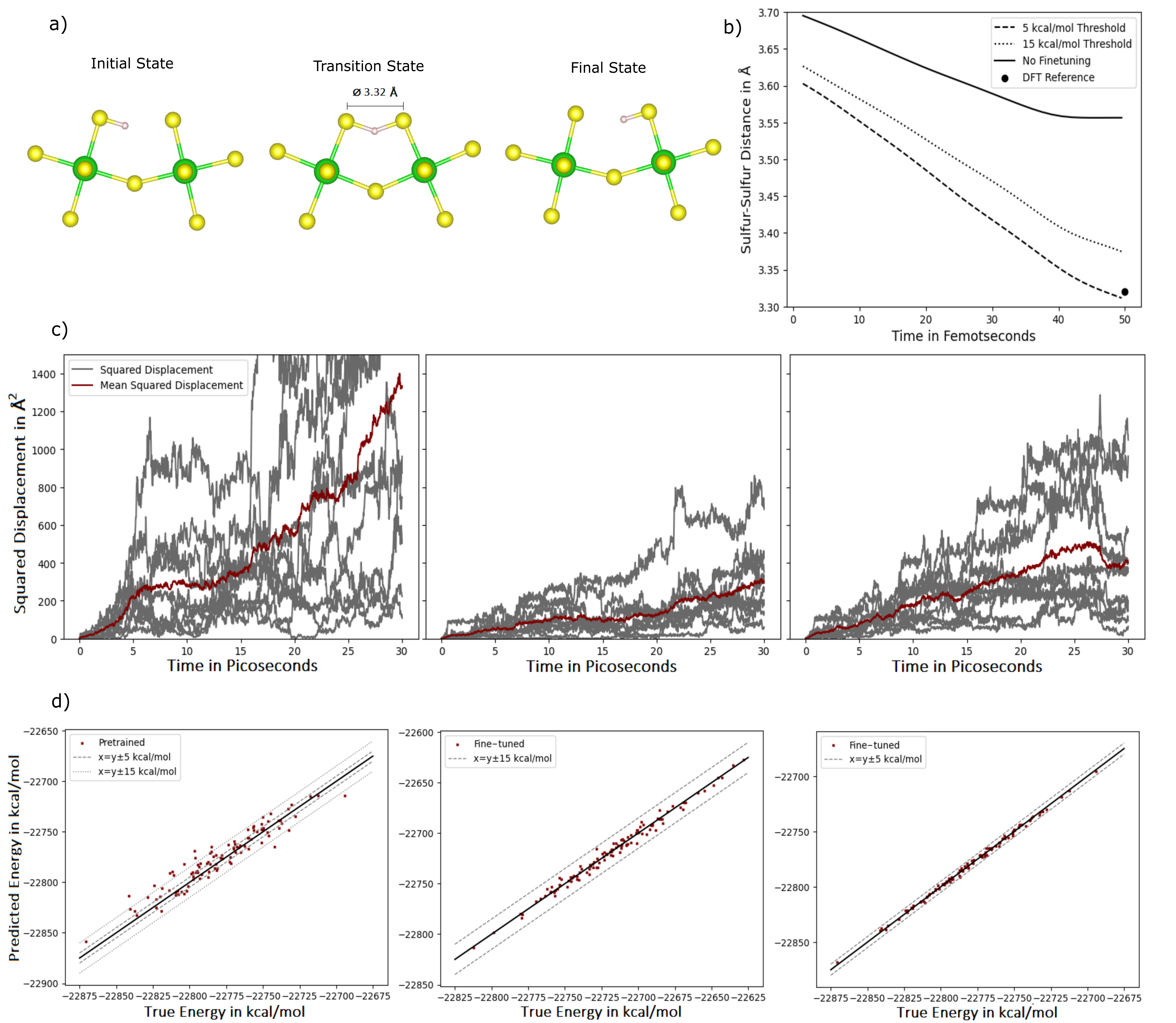}
\caption{a) An illustration of a proton jump between sulfur atoms in CaZrS\textsubscript{3}.\\
b) Average sulfur-sulfur distance in the 50 fs leading up to a proton jump for each of the models. The point at 50 fs indicates the average distance at the time of the jump computed from a 30-ps DFT trajectory in a previous research project.\\
c) Squared displacement from their starting position of the protons during the simulation. On the left, the non-fine-tuned model. In the middle the 15 kcal/mol threshold model. On the right, the 5 kcal/mol threshold model. The grey plots correspond to individual protons, while the red plots indicate the mean square displacement of all protons. \\
d) Errors of the models on the CaZrS\textsubscript{3} - proton system at randomly sampled time-steps. On the right, the 5 kcal/mol threshold model. In the middle, the 15 kcal/mol threshold model. On the left, the non-fine-tuned model.}\label{8}
\end{figure*}

Lastly, for the proton diffusion, we find that the models are outside the specified accuracy range of 5 and 15 kcal/mol at 2.5 and 3.2 percent of interventions, respectively. At interventions, the RMSEs of the models are 2.137 and 6.773 kcal/mol (0.57 and 1.78 meV/atom). For comparison, the non-fine-tuned model has an RMSE of 10.99 kcal/mol (2.91 meV/atom) on the intervention structures of the higher accuracy model. Towards the end of the training run, the disagreement thresholds for the models were 0.91 and 3.43 kcal/mol. This highlights the necessity for model calibration because the RMSE is about 2 times larger than the model disagreement. \\
After the initial training run, we use the fine-tuned models to analyse the Square Displacement of protons in CaZrS\textsubscript{3}. We perform five simulations with each model by first setting the target temperature with a Langevin thermostat during an initial one picosecond-long simulation. Afterwards, we perform a 30 ps simulation in the NVE ensemble. Because two protons were placed in the supercell of the simulation, this results in ten proton trajectories for each model.\\
The squared displacements for the different models are shown in Figure \ref{8} c. We see a very large Mean Square Displacement (MSD) for the non-fine-tuned model, which is reduced significantly when fine-tuned with an error threshold of 15 kcal/mol. Interestingly, fine-tuning to a lower error threshold of 5 kcal/mol leads to the MSD rising again slightly.\\
To explore the source of this phenomenon, we investigated the mechanism behind the proton jumps for the individual models. To do this, we analyzed the average sulfur-sulfur distance for the 50 fs leading up to proton jumps. For this, we have additional DFT reference data from a 30-ps DFT simulation from a previous research project in our group, where the average separation at the time of the jump was analyzed. The results are shown in  Figure \ref{8} b. For the low error threshold model, we find very good agreement with the DFT reference value. For the higher threshold model, the distance is slightly too large. This yields a possible explanation for the slightly lower proton mobility from this model, since the activation energy for jumps will be slightly larger at this separation. For the non-fine-tuned model, the average sulfur-sulfur distance at the time of the jump is much larger than the DFT reference. This demonstrates that the transition states underlying the proton jumps are not accurately modeled without finetuning, since the activation energy at such distances should be very high, resulting in low proton mobility instead of the erroneously high mobility that is predicted by the foundation model.\\
To further assess the accuracy of the models at test time, we sampled 100 random geometries from the first simulation for each of the already fine-tuned models. From the results shown in Figure \ref{8} d, it can be seen that the models are within their target accuracy for those configurations. For reference, we again include an evaluation of the error of the non-fine-tuned model on the geometries sampled from the simulation of the high-accuracy model and find a much poorer accuracy. \\
Overall, it took 31 DFT calls to train the model to the higher threshold and 285 calls to train it to the lower error threshold.\\
 We did an additional analysis to assess if, after the initial learning phase of 5000 simulation steps, there is a preference for DFT calls during transition states where the proton jumps between sulfur atoms. Interestingly, despite the large size of the system with 162 atoms, the overall uncertainty from the system did not completely drown out the uncertainty from the transition states in the low-threshold training run. In particular, we found a statistically highly significant increase in DFT calls (p = 0.0012) in 15 fs time windows around proton jumps, with an increase of 56 percent in the average number of DFT calls during time windows around proton jumps compared to time windows of equal size that do not contain proton jumps.
\section{Discussion}
As is illustrated on the NbSiAs surface - COH adsorbate benchmark dataset and in the appendix, the utilized transfer learning prior can improve the data efficiency substantially while enabling uncertainty-based detection of atomic structures that have an increased likelihood of a large prediction error. Furthermore, our full uncertainty-based on-the-fly fine-tuning workflow was able to maintain a specified accuracy and bias the training dataset toward rare events by selectively doing DFT reference calculations for samples with a high uncertainty. The workflow resulted in a significant improvement in the accuracy of the simulations over the non-fine-tuned models while only requiring a small amount of training data. The only simulation requiring more than 100 DFT interventions was for modeling the proton diffusion in CaZrS\textsubscript{3} to an exceptionally high accuracy, achieving an RMSE of 0.57 meV/atom at time of intervention and at a temperature of 1500 Kelvin. This increase in accuracy was particularly reflected in a much better modeling of the underlying physical processes and phenomena, such as the phase behaviour of LaMnO\textsubscript{3} due to subtle effects in the electronic structure at different temperatures and the transport mechanism of protons in CaZrS\textsubscript{3}.\\

Due to the simplicity of the uncertainty-aware transfer learning approach, it should be fairly simple to apply this approach to other neural network models not included in this work, since it worked almost out of the box for all neural network models investigated here. The only hyperparameters that we needed to fit for each model architecture individually were the strength of the BNN prior and the step size of the Markov chain. However, during our experiments, the same hyperparameters for each architecture gave good results across different simulations or datasets. This indicates that strong baseline values for those hyperparameters can be determined and given as default values to end-users. \\
In this work, we used the uncertainty in the energy predictions to decide if a DFT call should be done or not. An alternative approach would have been to use the uncertainty in the force predictions. We went with the energy predictions because accuracy in the energies controls the accuracy of the thermodynamic ensembles, and accurate forces do not necessarily guarantee accurate potential energies since even small errors in the forces can add up to large errors in the potential energy over large changes in the geometry. However, an advantage of using the force uncertainties would be that forces are localized. This makes it possible to choose individual error thresholds for each atom, where some atoms or regions might be of particular interest, such as in the case of the proton diffusion. For this reason, we plan to implement this in the future.\\

In regards to the speed up our fine-tuning approach can achieve for simulations, the exact results will be dependent on the chosen error threshold, hardware availability and the DFT software used. 
We give some illustrative examples and estimates for different hardware configurations for the CaZrS\textsubscript{3} - proton system in Appendix D.
Generally, it does not take a lot of training steps to update each Monte Carlo sample on a new DFT labeled geometry and we only do 2000 training steps at each iteration at a batch size of 5.
Additionally, updating the model is highly parallelizable because the Markov chain of each Monte Carlo sample can be run independently. 
 Nonetheless, on limited hardware, it might be better to only sample 4 instead of 8 Monte Carlo samples, for which we still found decent results in a previous work \cite{rensmeyer}. However, it is not advisable to go below 4 Monte Carlo samples as the quality of uncertainty quantification starts to degrade more significantly. In cases of very constrained GPU resources, using more computationally efficient variational inference-based BNN methods instead of MCMC methods might therefore be a better option. \\

Overall, the quality of uncertainty quantification was good enough to keep the model in the specified accuracy range and to bias the training set towards rare events. However, there were still instances where the error was large despite a small predicted standard deviation, which leaves room for improvement.\\
One possible cause for this might be that we sample the different sets of weights from a small region around the weights of the pre-trained model, because of the chosen prior over the weights. Recently, an approach to sampling Bayesian neural network weights has been introduced, where the prior can be specified on the function space instead of the weight space \cite{fBNN}. This is an attractive alternative because it would avoid biasing the weights of the model to stay close to the original ones and instead would only bias the weights to result in predictions that lie within a certain range of the original model. Attempting to adapt this formalism to construct a more sophisticated prior from the function space, based on the belief that the neural network might need to change its predictions by a certain amount to be accurate for the system of interest, might therefore be a promising approach to improve the quality of uncertainty quantification. \\
Other factors can additionally make uncertainty quantification challenging in an on-the-fly learning setting. 
There can be discontinuities in the electronic structure of molecules or materials, for example, when changes in the atomic geometry push previously occupied electronic states through the Fermi level.
This can cause discontinuities in the curvature of the potential energy surface, which makes uncertainty quantification challenging because current machine learning models are unaware of the electronic structure at this detail.  Because of this, the first occurrences of geometries with novel electronic structures in an on-the-fly learning scenario will likely always be predicted with some degree of overconfidence.\\
Lastly, we currently make the assumption in our calibration of the uncertainty that the miscalibration of the uncalibrated model uncertainties remains approximately constant over the course of the simulation. This assumption can, of course, be violated, for example, during phase transitions, and hence it might be beneficial to adjust the calibration procedure for such cases. One approach could be to not calculate $M_n$ from an average but instead from an exponential moving average that will weigh more recent examples more strongly compared to older ones.

\section{Conclusion}
In this work, we introduced an on-the-fly fine-tuning workflow for foundational neural network potentials that is capable of maintaining a user-specified accuracy by doing DFT reference calculations for high-uncertainty structures. As is illustrated in the paper and the appendix, the utilized transfer learning prior can improve the data efficiency substantially. Furthermore, the integration into our uncertainty-based on-the-fly fine-tuning workflow is very useful for biasing the fine-tuning dataset towards rare events and maintaining a user-specified accuracy while making the fine-tuning process easy to use for end-users, especially those without an extensive background in neural network potentials. Moreover, it enables an automated fine-tuning of foundation models within high-throughput materials discovery workflows. Given the broad applicability of the uncertainty-aware transfer learning approach across neural network architectures and the strong performance of the on-the-fly learning workflow, we believe this method has the potential to become a standard for fine-tuning foundational neural network potentials in machine learning–accelerated molecular dynamics simulations for materials science.  
 
\section{Acknowledgements}
This research as part of the project CouplteIT!
is funded by dtec.bw – Digitalization and Technology Research Center of the Bundeswehr which we gratefully acknowledge. dtec.bw is funded by the European Union –
NextGenerationEU.
Computational resources (HPC cluster HSUper) have been provided by the project hpc.bw, funded by dtec.bw.
\section{Code Availability}
The code for running on-the-fly fine-tuning simulations is available at \url{https://github.com/TimRensmeyer/OTFFineTune}.
% References
\bibliography{bib}

%%%END OF MAIN TEXT%%%


%\balance

%If notes are included in your references, you can change the title from 'References' to 'Notes and references' using the following command:

 %You need to replace "rsc" on this line with the name of your .bib file
%\bibliographystyle{ieee} %the RSC's .bst file
%\bibstyle{rsc}
\newpage
\appendix
\onecolumn

\section{Additional Transfer Learning Results}
In our initial investigation of the suitability of the chosen transfer learning prior, we did some additional transfer learning experiments with pre-trained NequIP models focused on the transfer learning of the forces. Due to the focus on the forces and the overall settings considered (e.g. transfer learning from lower to higher accuracy simulations), they do not fit well into the theme of the main manuscript. However, we believe that they may be of interest to other researchers so we included them here.

\subsection{Empirical Evaluation}
\begin{figure*}%[H]
\centering
\includegraphics[scale=0.73]{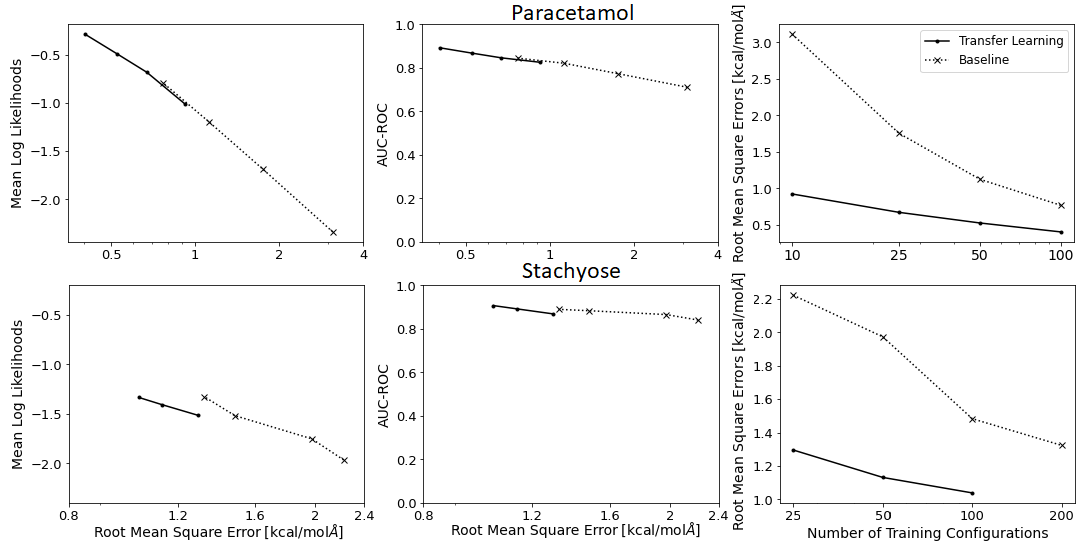}
\caption{Results on the paracetamol and stachyose datasets. On the left are the mean log-likelihoods as a function of RMSE (all means and standard deviations in kcal/mol\AA). In the middle are the AUC-ROC scores for uncertainty-based detection of force components with a high prediction error. On the right are the Root Mean Square Errors as a function of the number of training configurations}\label{A3}
\end{figure*}
We did three additional experiments to evaluate the transfer learning approach, representing likely scenarios where transfer learning might be employed. Notably, these experiments did not involve the on-the-fly learning active workflow but were instead done as tests of the prior over the parameters of the BNN on benchmark datasets.
The first test is a transfer learning scenario of finetuning a more general NequIP model trained on a variety of different compounds to a specific molecule of interest not included in the pre-training dataset. More specifically, we pre-train a NequIP model on a dataset consisting of a variety of compounds of the MD17 \citep{MD17} and MD22 \citep{MD22} datasets, which consist of MD trajectories of several molecules at DFT level accuracy, and then fine-tune it on the paracetamol dataset of the MD17 dataset. \\
The second benchmark is a transfer learning scenario from DFTB level accuracy to DFT level accuracy. In particular, we generate a large dataset of different configurations of a stachyose molecule in DFTB for pre-training and then utilize the stachyose data from the MD22 dataset for the transfer learning task.\\
The third test scenario is a transfer learning task for reaching CC level accuracy on an ethanol molecule starting from a model pre-trained on the corresponding ethanol data from the MD17 dataset. The CC-level dataset used for this was introduced by \citet{CCData}.\\

All experiments were done with 8 Monte Carlo samples generated from the same Markov chain, which has been identified as a good tradeoff between computational complexity and quality of uncertainty quantification in our previous work \citep{rensmeyer}. Details of all the datasets can be found in Appendix C. We set $\sigma_{TL}$ as $0.2$ for all experiments. 

\paragraph{
 The Evaluation Metrics:\\}

On all tasks, we evaluate the model's overall accuracy in terms of the Root Mean Square Error (RMSE) of the force components in dependence on the size of the training dataset.
%We analyze the accuracy of the transfer learned model in dependence of the size of the training dataset and compare it to a model that is trained from scratch with a simple Gaussian mean field prior $p(\boldsymbol{\theta})\sim N(\boldsymbol{0},I)$.\\
We analyze the transfer learning models accuracy and quality of uncertainty quantification in comparison to a model with a Gaussian mean field prior $p(\boldsymbol{\theta})\sim N(\boldsymbol{0},I)$. For the evaluation of the uncertainties, we compare the Mean Log Likelihoods (MLLs) of the force components or energies as a function of the RMSE for both models. To smooth each predicted distribution of the 8 Monte Carlo samples on this metric, we fit a normal distribution to the means and variances of each predicted distribution and use these smoothed distributions instead. Further, since the main goal of the uncertainty measure is the identification of configurations with a large error in the prediction, we evaluate the models in the task of detecting force components with a large prediction error based on the predicted uncertainty. More specifically, we analyze the corresponding AUC-ROC scores for detecting large errors via the predicted variance and plotting them as a function of the RMSE. On the ethanol and paracetamol datasets, errors of more than 1kcal/mol\AA\; were considered large, while on the more difficult stachyose dataset, the cutoff was set as 3kcal/mol\AA \; because an error of 1kcal/mol\AA \; could not be considered an outlier.
Since ethanol is a very small molecule, it was computationally feasible to include a deep ensemble containing 8 models trained from scratch as a second baseline (see Appendix B.6 for the training details).
\subsection{Results}
 \begin{figure*}%[H]
\centering
\includegraphics[scale=0.73]{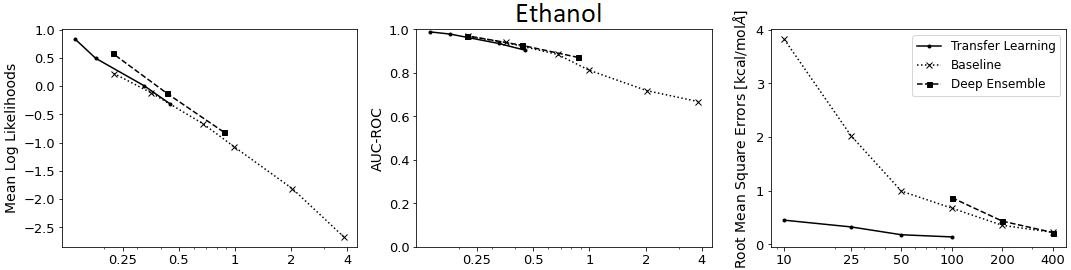}
\caption{Results on the ethanol dataset.  On the left are the mean log-likelihoods as a function of RMSE (all means and standard deviations in kcal/mol\AA). In the middle are the AUC-ROC scores for uncertainty-based detection of predictions with a large error. On the right are the Root Mean Square Errors as a function of the number of training configurations}\label{A4}
\end{figure*}
\begin{figure}[h]%[H]
%\centering
\includegraphics[scale=0.72]{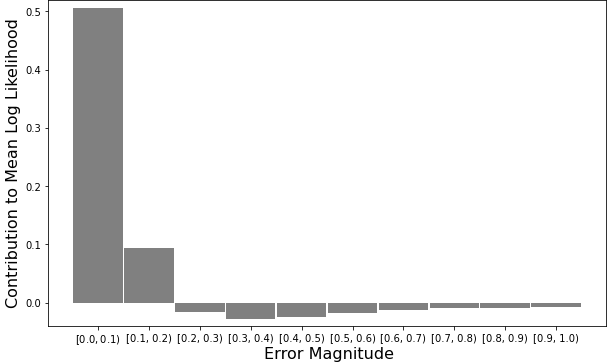}
\caption{Decomposition of the MLLs of the force components into contributions of different error magnitudes on the ethanol test set for the deep ensemble trained on 400 training configurations. The force units are in kcal/mol\AA.  }\label{A5}
\end{figure}

As can be seen in Figure \ref{A3}, very high accuracies were reached for the transfer learning model on the paracetamol dataset, even for small training datasets in terms of the RMSE when compared to the model trained from scratch. Further, there appears to be no major decrease in the quality of uncertainty quantification at a given accuracy as measured by the MLLs and AUC-ROC scores and the plots are almost on top of each other where the RMSEs overlap. However, there might be a very small decrease in quality as indicated by Figure \ref{A3}.\\
On the stachyose dataset, again, a clear improvement in accuracy at equal amounts of training samples is visible when compared to the baseline model (Figure \ref{A3}). However, both models have higher RMSEs than their counterparts on the paracetamol dataset at equal amounts of training configurations. The MLLs of the transfer learning model appear to be slightly lower than for a model trained from scratch when controlled for accuracy. The same is true only to a much smaller degree for the AUC-ROC scores. Further analysis revealed that the validation set was too small for the large configuration space of stachyose to properly recalibrate the uncertainties, which led to an overestimation of the errors on the test set for the transfer learned models but not for the baseline models. This also explains the absence of such reduced performance on the AUC-ROC scores, which are invariant under recalibration of uncertainties. Accounting for the slightly wrong calibration by recalibrating the uncertainties on the test set instead of the validation set confirmed miscalibration as the main source of the gap in MLLs. In particular, the gap between the MLLs of the transfer learning models that are closest in RMSE reduced from 0.191 to 0.086. Interestingly, we also attempted a transfer learning scenario on the stachyose dataset, where the model was only pre-trained on the small molecules of the MD17 dataset. However, this did not lead to any improvement in data efficiency. The most likely explanation for this is that the local atomic environments on the small molecule datasets, which the neural network uses to calculate potential energy contributions, are qualitatively very different from those of the stachyose molecule. \\
The biggest improvement in accuracy, when compared to the baseline model, was found on the ethanol dataset (Figure \ref{A4}), with an RMSE of less than 0.5kcal/mol\AA\; with only 10 configurations. There appears to be no decrease in the quality of uncertainty quantification, both in terms of MLLs as well as AUC-ROCs on this benchmark when compared to the baseline model. Comparing both the baseline model as well as the transfer learning model to the deep ensemble, almost identical performance can be seen on the outlier detection task, while the deep ensemble has slightly higher MLLs. A likely explanation for the selectively better performance in terms of MLLs but not on the outlier detection task is the fact that the MLLs are dominated by configurations with a smaller error, while, by definition, the outlier detection task requires good uncertainty quantification for samples with a larger prediction error. This can be illustrated by breaking down the MLLs into contributions from force components, whose prediction error falls into a certain interval, e.g. the contribution to the MLLs from samples where the prediction error is in the interval $[0.1,0.2)$ is given by the sum over the log-likelihoods of all force components in the test set where the prediction error is in the interval $[0.1,0.2)$ divided by the total number of force components in the test set. The results can be seen in Figure \ref{A5}. Consequently, it seems likely that deep ensembles achieve slightly better uncertainty quantification on configurations with a small prediction error but not on configurations with a large prediction error on this benchmark.\\

Interestingly, for all force transfer learning scenarios, the error of the pre-trained model was quite large with mean absolute errors of 2.31kcal/mol\AA \; on the paracetamol validation set, 4.34kcal/mol\AA \;on the stachyose validation set and 5.12kcal/mol\AA \;on the ethanol validation set. Further, all pre-trained models achieved a validation loss smaller than 0.15kcal/mol\AA \;on their pre-training datasets, strongly indicating that DFTB, DFT and CC methods disagree quite substantially in their force predictions for a given configuration. However, as was already alluded to in the introduction, this can most likely be traced back to simple biases in the simulation methods, such as slightly different equilibrium bond lengths. 
These small biases in different simulation methods can lead to qualitatively very similar force fields that may disagree substantially on the forces of a given configuration.
This would also explain why transfer learning is very efficient in these cases, as the model mostly has to correct for those biases, such as equilibrium bond lengths. 
Importantly, those force fields will lead to similar predictions of physical and chemical properties despite their apparently large disagreement, while a machine-learned force field with a similar magnitude of error to one of those methods can not in general be expected to yield those properties as well and hence needs to be trained to a much higher accuracy.\\
One additional result that stands out is the relatively high RMSE of both the transfer learning and the baseline model on the stachyose dataset when compared to the other two force-based test scenarios.
However, two factors make this dataset particularly challenging.
 First of all, stachyose is a larger molecule than paracetamol and ethanol, which in addition contains many single sigma bonds that allow for rotational degrees of freedom along the bond axis. This results in a very large configuration space for stachyose molecules, even relative to their size.
The second factor that makes this benchmark more challenging for the transfer learning model is that, unlike in the ethanol case, the higher accuracy dataset was not composed of configurations generated from an MD trajectory of the lower accuracy method but instead from a trajectory at DFT-level accuracy. As a result, the distribution of configurations in the DFT dataset will be different from the one from the DFTB dataset.\\
Lastly, one important observation we made is that the transfer learning approach converges much faster than when training from scratch. While state-of-the-art models can take days to train from scratch, training and validation losses converged within minutes on the transfer learning tasks. In fact, the only reason we let the sampling algorithm run for as long as described in Appendix B is to make sure that no pathological overfitting takes place.

\section{Implementation Details}
\subsection{Details of the Base Models}
 The foundation models in this work operate by first mapping the input 
$x=\{(\boldsymbol{r}_1, z_1), ...,(\boldsymbol{r}_n, z_n) \}$ and optionally the lattice vectors $\boldsymbol{L_1},\boldsymbol{L_2},\boldsymbol{L_3}$ to latent variables $\{\boldsymbol{v}_1, ..., \boldsymbol{v}_n\}$ that are invariant under distance-preserving transformations of the atomic coordinates $\boldsymbol{r}_i$. From those invariant atomic features, atomic energy contributions $\hat{E}_1, ..., \hat{E}_n$ are then calculated and summed up into a total potential energy prediction $\hat{E}=\sum_i E_i$. NequIP and MACE then calculate the forces acting on the atoms as the negative gradients of the potential energy $\hat{\boldsymbol{F}}_i=-\nabla_{\boldsymbol{r}_i}\hat{E}$ via automatic differentiation libraries while the Equiformerv2 model calculates the forces directly from a set of equivariant atomic feature vectors.
To apply the Bayesian neural network framework to these models, we modify the architectures slightly by adding layers, that compute standard deviations $\sigma_1, ..., \sigma_n$ for the forces from the invariant features $\{\boldsymbol{v}_1, ..., \boldsymbol{v}_n\}$.
Further, we introduce an energy standard deviation $\sigma_{\hat{E}}$ and model the distribution over the energy and forces as $$E,\boldsymbol{F}_1,..., \boldsymbol{F}_n |(\boldsymbol{r}_1, z_1), ...,(\boldsymbol{r}_n, z_n), \boldsymbol{\theta} \sim N(\hat{E},\sigma_{\hat{E}}^2)\Pi_{i=1}^n N(\hat{\boldsymbol{F}}_i,\sigma_i^2 I), $$
where $N$ denotes a normal distribution and $I$ is the identity matrix.\\
For the Equiformev2 and MACE models, the predictions are additionally conditioned on the lattice vectors $\boldsymbol{L_1},\boldsymbol{L_2},\boldsymbol{L_3}$.
For the MACE model, we also predict the stress tensor. For this, we modify the predicted distribution to
$$E,\boldsymbol{F}_1,..., \boldsymbol{F}_n ,\boldsymbol{S}|(\boldsymbol{r}_1, z_1), ...,(\boldsymbol{r}_n, z_n),\boldsymbol{L_1},\boldsymbol{L_2},\boldsymbol{L_3}, \boldsymbol{\theta} \sim N(\hat{E},\sigma_{\hat{E}}^2)N(\hat{\boldsymbol{S}},\sigma_{\hat{\boldsymbol{S}}}^2 I)\Pi_{i=1}^n N(\hat{\boldsymbol{F}}_i,\sigma_i^2 I), $$
where $\hat{\boldsymbol{S}}$ is a vector containing the predicted components of the stress tensor and $\sigma_{\hat{\boldsymbol{S}}}$ is a small fixed standard deviation that we set to $0.1/16$ kcal/mol$\text{\AA}^3$.
For the prior of the additional parameters from the added layers, we set the means to zero but use the same standard deviation as for the other parameters.\\

\subsection{The EquiformerV2-Based Neural Network Architecture}
 \begin{figure}[h!]
\centering
\includegraphics[scale=0.75]{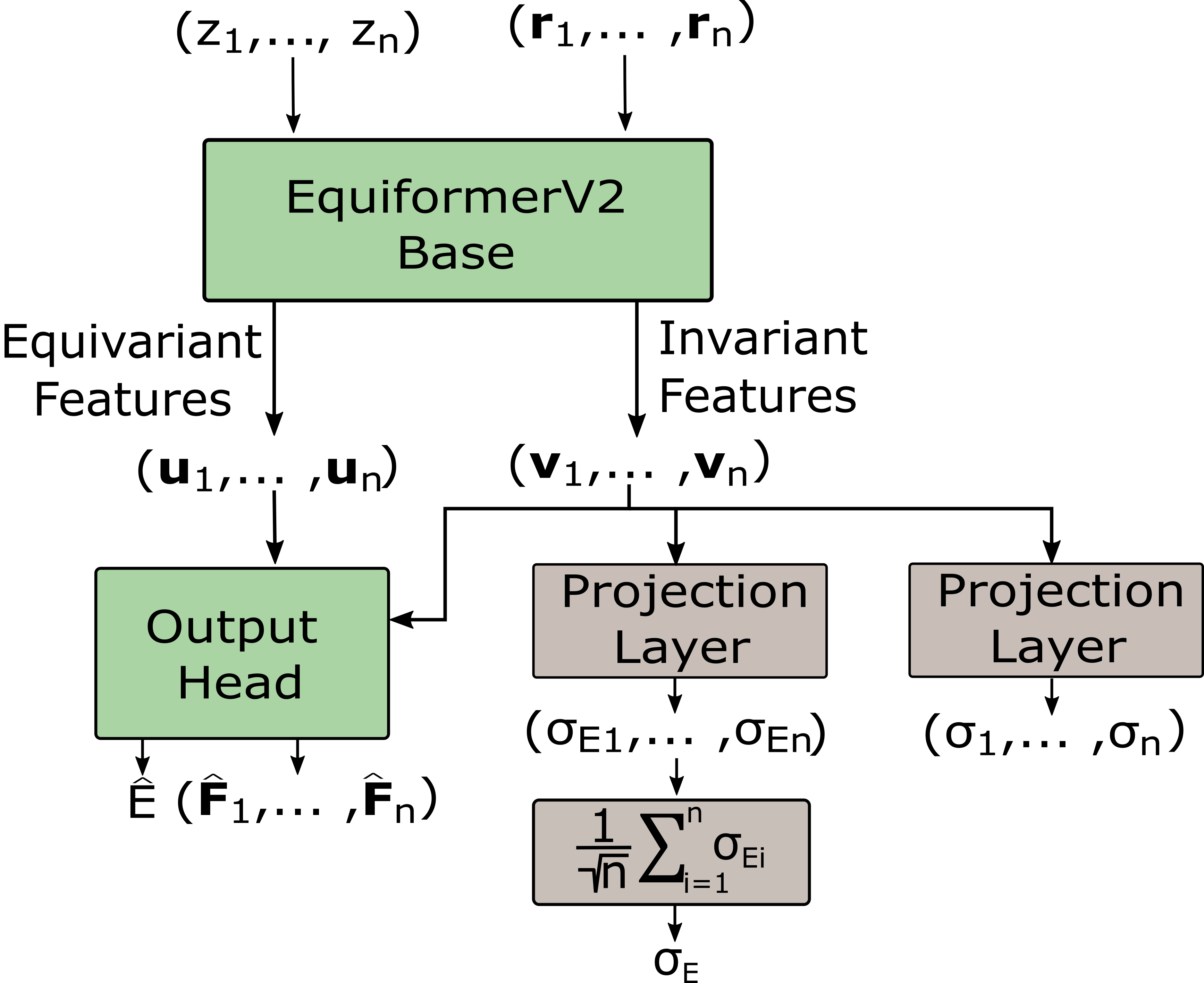}
\caption{The neural network architecture derived from the EquiformerV2 model used in the transfer learning task of potential energies. The green modules correspond to the modules of the EquiformerV2 architecture and the grey ones are additions to model the data probabilistically. The projection layers are simple linear layers followed by an exponential activation function.}\label{13}
\end{figure}
The overall architecture of the neural network derived from the EquiformerV2 model is summarized in Figure \ref{13}.
We chose the publicly available 31 million parameter EquiformerV2 model pre-trained on the entire OC20 dataset, including the MD data, as a base model for transfer learning.
Because this model is too large to train from scratch on such a relatively small training dataset, we chose a smaller EquiformerV2 model for the baseline model.
The configuration for that smaller model can be found on the first author's GitHub page.
Because the projection layers were not included in the pre-training, their means were set to zero in the transfer learning prior.

\subsection{The NequIP-Based Neural Network Architecture}

For the base neural network architecture, we use a NequIP model with four interaction blocks, a latent dimension of 64 and even and odd parity features up to and including angular momentum number l=2.\\
The standard deviations of the forces $\sigma_i$ are predicted by a three-layer MLP with input dimension 64, latent dimensions 32 and 16 and output dimension 1. \\ 
SiLU activation functions are used for the latent layers and the output activation function is the exponential function.\\
A base with 5 interaction blocks was used.\\
Because there aren't a lot of labeled examples for the potential energies, we just use a fixed energy standard deviation during the on-the-fly experiments, which we set to 10 percent of the target accuracy.\\
For the experiments in Appendix A, we don't include the energies in the training and train only on the forces.\\

\subsection{The MACE-Based Neural Network Architecture}
The modifications to the MACE model follow the same pattern as for the NequIP model. \\
The output of the mace-mp medium model contains feature vectors for each atom. 256 components of each feature vector are rotation invariant. From these features, the standard deviations of the forces $\sigma_i$ are predicted by a three-layer MLP with input dimension 256, latent dimensions 64 and 16 and output dimension 1. \\
We again use a fixed energy standard deviation set to 10 percent of the target accuracy. For all experiments, we further use a fixed stress standard deviation of $0.1/16$ kcal/mol$\text{\AA}^3$.

\subsection{Generating Samples from the Posterior}
For sampling the Bayesian posterior, we use the SGHMC algorithm \citep{SGHMC} with the adaptive mass term introduced by us in a previous work \citep{rensmeyer}.
\paragraph{Sampling for the Transfer Learning Experiments:\\}
For the ethanol and paracetamol test cases, the step size $\gamma$ is exponentially decreased from $10^{-2}$ and $0.3\cdot10^{-2}$ to $10^{-5}$ during the first $10^6$ steps for the baseline model and transfer learning model, respectively. At the end of this phase, the first model is sampled. Afterward, we use the cyclical learning rate schedule:\\
$$\gamma_i = \frac{\gamma_0}{2}\left(\cos\left(\pi+\frac{i\cdot \pi}{K}\right)+1\right)$$ with $\gamma_0=0.001$ and cycle length $K=50000$\\
 to sample the subsequent models from the same Markov chain at the end of each cycle.\\
The same procedure is also utilized for the baseline model on the stachyose test case. However, the initial convergence phase is shortened to $0.5\cdot 10^6$ steps for the transfer learning model, as the other two test cases had revealed a quicker convergence for the transfer learning models.
For the paracetamol and ethanol cases, a batch size of 30 is used and for the stachyose case, it is set as 15.
For the transfer learning experiments with the NequIP model, we only trained on the forces, while for the Equiformerv2 model, we trained on both forces and energies. Analogous to the on-the-fly learning experiments, we calculate an energy offset for the training, validation and test data, so that for the first sample in the training data, the energy equals the one predicted by the pre-trained model.
For the surface-adsorbate transfer learning task, a batch size of 20 was used for the baseline model and for the transfer learning task, it is set as 15.
Furthermore, the same sampling procedure is employed. For the baseline and transfer learning model, the step size $\gamma$ is exponentially decreased from $10^{-4}$ and $\cdot10^{-5}$ to $10^{-7}$ and $10^{-8}$ respectively during the initial convergence phase. This phase was $0.5 \cdot 10^6$ steps long for the baseline model and $10^5$ steps long for the transfer learning model, respectively.
Then again, a cyclical sampling procedure is employed to generate the other samples with $\gamma_0=0.0001$ and cycle length $K=50000$ 
After the first $90$ percent of the initial convergence phase, the mass term is kept constant to ensure close convergence to the posterior.

\paragraph{Sampling for the on-the-fly Finetuning Experiments:\\}
To sample the posterior for the on-the-fly fine-tuning experiments, we run 8 separate Markov Chains, one for each Monte Carlo sample. After each added training sample, we run the SGHMC algorithm with the adaptive mass term for 2000 steps to update the Monte Carlo samples. We use a batch size of 5 and a learning rate schedule 
$$\gamma_i = \frac{\gamma_0}{2}\left(\cos\left(\frac{i\cdot \pi}{2000}\right)+1\right).$$
For the MACE model, we use $\gamma_0=0.001$ and for the NequIP model, we set $\gamma_0=0.00003$.
To improve the speed of convergence, we use priority sampling to increase the likelihood of sampling the newly added training sample at each iteration. We adjust the sampling probability so that, on average, the newly added sample is contained once in each minibatch. We weigh each sample in the minibatch estimator of the log-likelihood by the likelihood ratio of a uniform sampling procedure and the sampling probabilities used to ensure the minibatch estimator is unbiased.

\subsection{Details on the Simulations}
\paragraph{The Ethanol On-The-Fly Simulation:}
We used the Langevin thermostat with 0.5 fs time steps and a friction term of $0.01 fs^{-1}$ of the Atomic Simulation Environment (ASE) library to drive the dynamics. The DFT calculations were done with the Vienna Ab initio Simulation Package (VASP) using a cubic 30 \AA \;simulation cell. We used a plane wave energy cutoff of 800 eV with a convergence criterion of $1e-6$ eV and the B3LYP exchange-correlation functional \cite{B3LYP1,B3LYP2,B3LYP3,B3LYP4}.
\paragraph{The LaMnO\textsubscript{3} On-The-Fly Simulation:}
We used the NPT thermostat of the Atomic Simulation Environment (ASE) library with a 0.5 fs time step, a ttime of 100 fs, a pfactor of $160 GPa\cdot 0.1\cdot 75^2 fs^2$ and an external pressure of 1 bar to drive the dynamics. We use a 2x2x2 supercell containing 40 atoms in total. The DFT calculations were done with the Vienna Ab initio Simulation Package (VASP). We used a plane wave energy cutoff of 500 eV with a convergence criterion of $1e-4$ eV and the PBE exchange-correlation functional \cite{PBE}. We used a Hubbard term of $3.9$ for the Mn atoms.
\paragraph{The Proton Diffusion On-The-Fly Simulations:}
We used the Langevin thermostat with a friction term of 0.5 of the Atomic Simulation Environment (ASE) library to drive the dynamics for 1 ps to set the temperature.
Afterwards, we used the velocity Verlet integrator to continue the simulation for another 30 ps. One initial training run was done for each of the two fine-tuned models. Afterwards, 5 production runs were done to investigate the diffusivity of the protons. 1.5 fs time steps were used for all simulations.
 We used a CaZrS\textsubscript{3} supercell containing 160 atoms and two additional protons. The DFT calculations were done with the Vienna Ab initio Simulation Package (VASP). We used a plane wave energy cutoff of 510 eV with a convergence criterion of $1e-4$ eV and the PBE exchange-correlation functional \cite{PBE}.
\subsection{Pretraining the Models}
\paragraph{Pretraining for the Transfer Learning Experiments:\\}
To pre-train a model, we converge it to a local maximum of the log-posterior on the pre-training dataset with a Gaussian mean field prior $p(\boldsymbol{\theta})\sim N(\boldsymbol{0},I)$. Almost the same sampling algorithm and hyperparameters are used as in the sampling of the posterior of the corresponding baseline model. The only differences are that the injected noise is downscaled by a factor of 0.1 and only the first model is sampled. The injected noise was not set to zero, because we found that a small amount of injected noise actually speeds up convergence, especially at the beginning of the optimization.
\paragraph{Pretraining the NequIP Model on the Spice Dataset for the Ethanol On-The-Fly experiment:\\}
We train the NequIP model at a batch size of 25 at a fixed learning rate of 3e-5 with the Adam optimizer. We use the loss function $L=(1/200)MSE(Energy) + MSE(Forces)$, where the Mean Square Error (MSE) refers to the batch mean. We keep a validation set of 70 structures from the SPICE dataset, which are not included in the training dataset. Every 20000 training steps, we evaluate the energy MSE on the validation set. We stop the training after the energy validation loss hasn't improved for 3 epochs. The final model used is the one with the lowest validation loss during the training run.
\subsection{Training and Evaluating the Deep Ensemble}
To generate the deep ensemble, we train 8 stochastic NequIP models from scratch with different random initializations of the neural network parameters. The models are trained with the AMSGrad optimizer at a batch size of 30 with an initial learning rate of 0.01, which is decayed to $10^{-5}$ over the course of $5\cdot 10^5$ training steps. Every $1000$ training steps, the model's RMSE is evaluated on a validation set of size 10. The parameter set with the best RMSE during the optimization procedure for each weight initialization is used to make predictions on the test set. We again fit a normal distribution to the predictions of the ensemble and recalibrate those uncertainties on the validation set when evaluating the MLLs.

\subsection{Calculation of Densities During Inference}
To smooth the predicted distribution of several Monte Carlo samples or ensemble models, we calculate the final distribution by fitting a normal distribution to the predicted means and variances.
The total variance of several Monte Carlo samples or Ensemble models for force components was calculated as 
$$\sigma_{F_i}^2=Variance\left(\hat{F}_{i,j}\right)+\frac{1}{k}\sum_{j=1}^k\sigma_{F_{i,j}}^2,$$
where $j$ enumerates the predicted standard deviations of the individual Monte Carlo samples/ ensemble models, $\hat{F}_{i,j}$ is the predicted expectation value for the i-th force component of that particular model and $\sigma_{F_{i,j}}$ the corresponding standard deviation. The variance is calculated over the Monte Carlo samples/ ensemble models.
The mean of the predicted distribution was simply calculated as $\hat{F}_{i}=\frac{1}{k}\sum_{j=1}^k \hat{F}_{i,j}$.\\
The calculation of the final energy and stress distribution was done completely analogously.

\subsection{Construction of the Estimator for the Relationship between Error and Predicted Standard Deviation from Figure \ref{8}.}
To construct this estimator, we ordered the pairs of predicted standard deviations and observed errors $\{(\sigma_1,e_1),...,(\sigma_{1800},e_{1800})\}$ by the magnitude of the predicted standard deviation. We then applied a Gaussian filter with a sigma value of 200 to both the list of ordered variances  $[\sigma_{sorted,1}^2,...,\sigma_{sorted,1800}^2]$ and squared errors $[e_{sorted,1}^2,...,e_{sorted,1800}^2]$ resulting in the smoothed arrays $[\sigma_{smoothed,1}^2,...,\sigma_{smoothed,1800}^2]$ and $[e_{smoothed,1}^2,...,e_{smoothed,1800}^2]$. Finally, we plot the root of the smoothed predicted variances over the root of the smoothed squared errors to generate the figure.

\subsection{The Bayesian Calibration Estimator}
Given a dataset of empirical observations of (independent) errors $E=\{e_1,...,e_n\}$ and predicted uncertainties $\Sigma=\{\sigma_{1},...,\sigma_{n}\}$, the error $e^*$ on a new sample with predicted standard deviation $\sigma^*$ can be given in closed form as the students t-distribution
$$p(e^*|\sigma^*,E,\Sigma)=\int p(e^*|\sigma^*,\lambda)p(\lambda|E,\Sigma)d\lambda$$
$$=\frac{1}{\int p(E,\Sigma|\lambda)p(\lambda)d\lambda}\cdot\int p(e^*|\sigma^*,\lambda)p(E,\Sigma|\lambda)p(\lambda)d\lambda$$
$$=\frac{\Gamma(a+\frac{n+1}{2})}{\sqrt{2\pi\sigma^{*2}}\Gamma(a+\frac{n}{2})}\cdot\frac{(b+\frac{1}{2}n\cdot M_n)^{a+\frac{n}{2}}}{(b+\frac{1}{2}n\cdot M_n+\frac{1}{2}\frac{e^{*2}}{\sigma^{*2}})^{a+\frac{n+1}{2}}},$$
where $M_n=\frac{1}{n}\sum_{i=1}^n\frac{e_i^2}{\sigma_i^2}$.
By integrating this density from $e^*=-K$ to $e^*=K$, the result:
$$p(|e^*|<K|\sigma^*,E,\Sigma)=\frac{2K\Gamma(a+\frac{n+1}{2})}{\sqrt{2\pi\sigma^{*2}}\Gamma(a+\frac{n}{2})\sqrt{b+\frac{1}{2}n\cdot M_n}}$$
$$\;\;\;\;\;\;\; \times Hyp2F1\left(\frac{1}{2},a+\frac{n+1}{2};\frac{3}{2},-\frac{K^2}{\sigma^{*2}(2b+n\cdot M_n)}\right),$$
can be derived.
Further, with the identities $\frac{\Gamma(x+\frac{1}{2})}{\Gamma(x)\sqrt{x}}\rightarrow 1$ and $(1+\frac{c}{x})^x \rightarrow e^c$ for $x \rightarrow \infty$ it can be verified, that for $n\rightarrow \infty$ the predicted error distribution becomes 
$$p(e^*|\sigma^*,E,\Sigma) \sim \mathcal{N}\left(0,\frac{\sigma^{*2}}{M_n}\right).$$

\section{The Datasets}
\subsection{The Ethanol Transfer Learning Datasets}
To pre-train the model, 5000 randomly sampled configurations from the MD17 ethanol dataset are used. This dataset consists of over 500000 configurations generated from a molecular dynamics trajectory calculated at DFT level accuracy.
We use the training and test datasets of ethanol at CCSD(T) level accuracy introduced by \citet{CCData} for the transfer learning task. We use the last 10 configurations of the training set as validation data. The actual training data consisted of the first $m \in \mathbb{N}$ configurations of the training dataset for varying values of $m$.
\subsection{The Paracetamol Transfer Learning Datasets}
The pretraining dataset consists of randomly sampled configurations from the aspirin, benzene, malonaldehyde, toluene, salicylic acid, naphthalene, ethanol, uracil and azobenzene from the MD17 dataset, as well as the AT-AT DNA base pair, stachyose, Ac-Ala3-NHMe, and docosahexaenoic acid datasets from the MD22 dataset. The first 100000 configurations from each MD17 dataset and all configurations from the MD22 datasets were used to form a pool of configurations from which 100000 are randomly drawn as the pretraining dataset.\\
For the actual training set $m \in \mathbb{N}$, configurations are randomly sampled from the MD17 paracetamol dataset for varying values of $m$. 10 additional configurations are randomly sampled as a validation set. The rest of the 
106490 configurations are used as a test set.
\subsection{The Stachyose Transfer Learning Datasets}

The pretraining dataset was generated from a long molecular dynamics trajectory of a stachyose molecule in DFTB+\citep{DFTB+}. The initial geometry was generated from a structural relaxation with a convergence criterion of 10$^{-3} H/$\AA \; for the maximal force component. The MD trajectory was simulated at 1 femtosecond time steps with a Nose Hoover thermostat \citep{NH} at 600° Kelvin with a coupling strength of 3200 cm$^{-1}$. The simulation ran for $10^6$ time steps using the velocity verlet driver with one configuration sampled every ten time steps, yielding a dataset of 100000 configurations. For both the geometry optimization as well as the MD simulation, a Hamiltonian with self-consistent charges \citep{SCC} and third-order corrections \citep{DFTB3}
 was used in correspondence with the 3ob-3-1 Slater Coster files \citep{3OB}. For all atoms, s- and p-orbitals were used in the Hamiltonian.\\ \\

For the actual training set $m \in \mathbb{N}$, configurations are randomly sampled from the first 10000 configurations of the MD22 stachyose dataset for varying values of $m$. 10 additional configurations are randomly sampled from the configurations 10100 to 10900 as a validation set. Configurations 11000 up to 27000 are used as a test set.

\subsection{The Surface-Adsorbate Dataset}
For the energy transfer learning dataset of the surface adsorbate system, we chose the NbSiAs surface - COH adsorbate dataset from the OC20-Dense dataset.\\
We randomly sampled 8000 configurations as possible training configurations. For a training dataset of size $n\in \mathbb{N}$, we chose the first $n$ configurations from that subset as a training set. We further randomly sampled twenty configurations as a validation set to recalibrate the uncertainties. The rest of the configurations were used as the test set.

\section{Runtime Estimates on Different Hardware Configurations for the CaZrS\textsubscript{3} - Proton System}
A single 30-ps proton diffusion simulation in CaZrS\textsubscript{3} would take around 3 months on the two Intel Platinum 8360Y processors we used to do the DFT interventions in VASP, while it took less than a day to finetune the 15 kcal/mol threshold model and around a week for the 5 kcal/mol threshold model during the initial 30-ps training runs. The time for the 5 production runs with each model was negligible. Increasing the CPU resources to eight processors will reduce the time of the simulation in VASP to around one month and also reduce the time for the on-the-fly simulations to less than half of the previous values, since the DFT interventions were the computational bottleneck. Increasing CPU resources even more will start to result in diminishing returns, as the MPI communications overhead will start to become the limiting factor. \\
We used GPU nodes containing 8 L40S GPUs for our experiments and updating all Monte Carlo samples on a new training data point takes only a few minutes on that hardware.
To test the impact of constrained GPU resources, we reran the experiment for the largest system, the 162-atom CaZrS\textsubscript{3}-proton system, using only two A100 GPUs. On that hardware, it takes around 20 minutes to update the model. However, it should be noted that DFT calculations for systems of that size will also be quite time-intensive and on the two Intel Platinum 8360Y processors we used, they took around 24-25 minutes.

\end{document}